# Reasoning within Fuzzy Description Logics

**Umberto Straccia**                                        STRACCIA@IEI.PI.CNR.IT
*I.E.I - C.N.R., Via G. Moruzzi, 1*
*I-56124 Pisa (PI), ITALY*

## Abstract

Description Logics (DLs) are suitable, well-known, logics for managing structured knowledge. They allow reasoning about individuals and well defined concepts, *i.e.* set of individuals with common properties. The experience in using DLs in applications has shown that in many cases we would like to extend their capabilities. In particular, their use in the context of Multimedia Information Retrieval (MIR) leads to the convincement that such DLs should allow the treatment of the inherent imprecision in multimedia object content representation and retrieval.

In this paper we will present a fuzzy extension of $\mathcal{ALC}$, combining Zadeh's fuzzy logic with a classical DL. In particular, concepts becomes fuzzy and, thus, reasoning about imprecise concepts is supported. We will define its syntax, its semantics, describe its properties and present a constraint propagation calculus for reasoning in it.

## 1. Introduction

The representation of uncertainty and imprecision has received a considerable attention in the Artificial Intelligence community in an attempt to extend existing knowledge representation systems to deal with the imperfect nature of real world information (which is likely the rule rather than an exception). An impressive work has been carried out in the last decades, resulting in a number of concepts being investigated, a number of problems being identified and a number of solutions being developed (Bacchus, 1990; Dubois & Prade, 1996; Kruse, Schwecke, & Heinsohn, 1991; Pearl, 1988).

For most knowledge representation formalisms, First-Order Logic (FOL) has been the basis: its basic units –individuals, their properties, and the relationship between them– naturally capture the way in which people encode their knowledge. Unfortunately, it is severely limited both (*i*) by its ability to represent our uncertainty about the world –due to lack of knowledge about the real world a fact can only estimated to be true to *e.g.* a probability degree; and (*ii*) by its ability to represent inherently imprecise knowledge– indeed, there are concepts, like hot, for which no exact definition exists and, thus, a fact like "35° Celsius is hot", rather being true or false, has a truth-value in between true and false.

In the last decade a substantial amount of work has been carried out in the context of *Description Logics* (DLs).[1] DLs are a logical reconstruction of the so-called frame-based knowledge representation languages, with the aim of providing a simple well-established Tarski-style declarative semantics to capture the meaning of the most popular features of structured representation of knowledge. A main point is that DLs are considered as to be attractive logics in knowledge based applications as they are a good compromise between expressive power and computational complexity.

---

1. Description Logics have also been referred to as Terminological Logics, Concept Logics, KL-ONE-like languages. The web page of the description logic community is found at address `http://dl.kr.org/dl`.

Nowadays, a whole family of knowledge representation systems has been build using DLs, which differ with respect to their expressiveness, their complexity and the completeness of their algorithms, and they have been used for building a variety of applications (Peltason, 1991; Brachman, 1992; Baader & Hollunder, 1991a; Horrocks, 1998).

Experience in using DLs in applications has also shown that in many cases we would like to extend the representational and reasoning capabilities of them. In particular, the use of DLs in the context of Multimedia Information Retrieval (MIR) points out the necessity of extending DLs with capabilities which allow the treatment of the inherent imprecision in multimedia object representation and retrieval (Meghini & Straccia, 1996; Meghini, Sebastiani, & Straccia, 1997, 1998). In fact, classical DLs are insufficient for describing *real* multimedia retrieval situations, as the retrieval is usually not only a yes-no question: (*i*) the representations of multimedia objects' content and queries which the system (and the logic) have access to are inherently imperfect; and (*ii*) the relevance of a multimedia object to a query can thus be established only up to a limited degree. Because of this, we need a logic in which, rather than deciding *tout court* whether a multimedia object satisfies a query or not, we are able to *rank* the retrieved objects according to how strongly the system believes in their relevance to a query.

To this end, we will extend DLs with *fuzzy* capabilities. The choice of fuzzy set theory as a way of endowing a DL with the capability to deal with imprecision is not uncommon (da Silva, Pereira, & Netto, 1994; Tresp & Molitor, 1998; Yen, 1991) and can be motivated

- from a semantics point of view, as fuzzy logics capture the notion of imprecise concept, *i.e.* a concept for which a clear and precise definition is not possible. Fuzzy concepts play a key role in *e.g.* content descriptions of multimedia objects (most of human's concepts are imprecise). For instance, in the context of images, the (semantic) content of an image region $r$ may be described by means of a fuzzy statement like "$r$ is about a Ferrari" and establish that this sentence has truth-value 0.8, *i.e.* $r$ is likely about a Ferrari;

- from a proof theoretical point of view, as there exist well-known techniques for reasoning in fuzzy logics (Chen & Kundu, 1996; Lee, 1972; Xiachun, Yunfei, & Xuhua, 1995).

In the following we will present a quite general fuzzy DL, in the sense that it is based on the DL $\mathcal{ALC}$, a significant and expressive representative of the various DLs. This allows us to adapt it easily to the different DLs presented in the literature. Another important point is that we will show that the additional expressive power has no impact from a computational complexity point of view. This is certainly important as the nice trade-off between computational complexity and expressive power of DLs contributes to their popularity.

Note that our fuzzy extension for the management of imprecise knowledge is complementary to other DL extensions for the management of uncertainty, *e.g.* probabilistic extension (Heinsohn, 1994; Jäger, 1994; Koller, Levy, & Pfeffer, 1997; Sebastiani, 1994) with some exceptions like shown by Hollunder (1994) where a possibilistic DL has been considered. Even though these probabilistic extensions enlarge the applicability of DLs they do not address the issue of reasoning about individuals and imprecise concepts, as imprecise knowledge and uncertain knowledge are "orthogonal" (Dubois & Prade, 1994). Moreover, reasoning in a probabilistic framework is generally a harder task, from a computational point of view, than the relative non probabilistic case and in most cases a complete axiomatization is missing (Halpern, 1990; Roth, 1996). As a consequence, the computational problems have to be addressed carefully (Koller et al., 1997).

We will proceed as follows. In the following section we first introduce $\mathcal{ALC}$. In Section 3 we extend $\mathcal{ALC}$ to the fuzzy case and discuss some properties in Section 4, while in Section 5 we will present a constraint propagation calculus for reasoning in it. Section 6 concludes and presents some topics for further research.

## 2. A Quick Look to $\mathcal{ALC}$

The specific DL we will extend with fuzzy capabilities is $\mathcal{ALC}$, a significant representative of DLs. At first, we will introduce classical $\mathcal{ALC}$, while in Section 3 our fuzzy extension of $\mathcal{ALC}$ will be presented.

We assume three alphabets of symbols, called *primitive concepts* (denoted by $A$), *primitive roles* (denoted by $R$) and *individuals* (denoted by $a$ and $b$).[2]

### 2.1 Concept and Role

Concepts are expressions that collect the properties, described by means of roles, of a set of individuals. From a FOL point of view, concepts can be seen as unary predicates, whereas roles are interpreted as binary predicates.

A *concept* (denoted by $C$ or $D$) of the language $\mathcal{ALC}$ is build out of primitive concepts according to the following syntax rules:

$$
\begin{array}{rcll}
C, D & \longrightarrow & \top \mid & \text{(top concept)} \\
& & \bot \mid & \text{(bottom concept)} \\
& & A \mid & \text{(primitive concept)} \\
& & C \sqcap D \mid & \text{(concept conjunction)} \\
& & C \sqcup D \mid & \text{(concept disjunction)} \\
& & \neg C \mid & \text{(concept negation)} \\
& & \forall R.C \mid & \text{(universal quantification)} \\
& & \exists R.C & \text{(existential quantification)}.
\end{array}
$$

### 2.2 Interpretation

DLs have a clean, model-theoretic semantics, based on the notion of interpretation. An *interpretation* $\mathcal{I}$ is a pair $\mathcal{I} = (\Delta^{\mathcal{I}}, \cdot^{\mathcal{I}})$ consisting of a non empty set $\Delta^{\mathcal{I}}$ (called the *domain*) and of an *interpretation function* $\cdot^{\mathcal{I}}$ mapping different individuals into different elements of $\Delta^{\mathcal{I}}$ (called *unique name assumption*), primitive concepts into subsets of $\Delta^{\mathcal{I}}$ and primitive roles into subsets of $\Delta^{\mathcal{I}} \times \Delta^{\mathcal{I}}$. The interpretation of complex concepts is defined as usual:

---

2. Through this work we assume that every metavariable has an optional subscript or superscript.

$$
\begin{aligned}
\top^{\mathcal{I}} &= \Delta^{\mathcal{I}} \\
\bot^{\mathcal{I}} &= \emptyset \\
(C \sqcap D)^{\mathcal{I}} &= C^{\mathcal{I}} \cap D^{\mathcal{I}} \\
(C \sqcup D)^{\mathcal{I}} &= C^{\mathcal{I}} \cup D^{\mathcal{I}} \\
(\neg C)^{\mathcal{I}} &= \Delta^{\mathcal{I}} \setminus C^{\mathcal{I}} \\
(\forall R.C)^{\mathcal{I}} &= \{d \in \Delta^{\mathcal{I}} : \forall d'.(d,d') \notin R^{\mathcal{I}} \text{ or } d' \in C^{\mathcal{I}}\} \\
(\exists R.C)^{\mathcal{I}} &= \{d \in \Delta^{\mathcal{I}} : \exists d'.(d,d') \in R^{\mathcal{I}} \text{ and } d' \in C^{\mathcal{I}}\}.
\end{aligned}
$$

Note that each concept $C$ and role $R$ can be mapped into an equivalent open first-order formula $F_C(x)$ and $F_R(x,y)$, respectively:

$$
\begin{aligned}
F_\top(x) &= \mathsf{T} & (1) \\
F_\bot(x) &= \mathsf{F} & (2) \\
F_A(x) &= A(x) & (3) \\
F_R(x,y) &= R(x,y) & (4) \\
F_{C \sqcap D}(x) &= F_C(x) \wedge F_D(x) & (5) \\
F_{C \sqcup D}(x) &= F_C(x) \vee F_D(x) & (6) \\
F_{\neg C}(x) &= \neg F_C(x) & (7) \\
F_{\forall R.C}(x) &= \forall y. \neg F_R(x,y) \vee F_C(y) & (8) \\
F_{\exists R.C}(x) &= \exists y. F_R(x,y) \wedge F_C(y), & (9)
\end{aligned}
$$

where $\mathsf{T}$ and $\mathsf{F}$ are two formulae representing the truth-value "true" and "false", respectively (*e.g.* $\mathsf{T} = p \vee \neg p$ and $\mathsf{F} = p \wedge \neg p$, for some letter $p$).

Two concepts $C$ and $D$ are said to be *equivalent* (denoted by $C \equiv D$) when $C^{\mathcal{I}} = D^{\mathcal{I}}$ for all interpretations $\mathcal{I}$. Note that, *e.g.* $\top \equiv \neg \bot$; $C \sqcap D \equiv \neg(\neg C \sqcup \neg D)$, and $(\exists R.C) \equiv (\neg \forall R. \neg C)$.

### 2.3 Assertion

An *assertion* (denoted by $\alpha$) is an expression of type $a{:}C$ ("$a$ is $C$", also "$a$ is an instance of $C$"), or an expression of type $(a,b){:}R$ ("$(a,b)$ is $R$", also "$(a,b)$ is an instance of $R$"). For instance, $\mathsf{tom{:}Tall \sqcap Student}$ asserts that Tom is a tall student, whereas $(\mathsf{tim, tom}){:}\mathsf{Friend}$ asserts that Tom is a friend of Tim. A *primitive assertion* is either an assertion of the form $a{:}A$, where $A$ is a primitive concept, or an assertion of the form $(a,b){:}R$. From a semantics point of view, an interpretation $\mathcal{I}$ *satisfies* $a{:}C$ (resp. $(a,b){:}R$) iff $a^{\mathcal{I}} \in C^{\mathcal{I}}$ (resp. $(a^{\mathcal{I}}, b^{\mathcal{I}}) \in R^{\mathcal{I}}$).

### 2.4 Terminological Axiom

A *terminological axiom* (denoted by $\tau$) is either a concept specialisation or a concept definition. A *concept specialisation* is an expression of the form $A{<}C$, where $A$ is a primitive concept and $C$ is a concept. A specialisation allows stating the existence of a specialisation ("more specific than") relation between concepts. For instance, $\mathsf{Ferrari{<}SportCar \sqcap}$ $\exists$Ownedby.CarFanatic states that a Ferrari is a sport car that is owned by a car fanatic. On the other hand, a *concept definition* is an expression of the form $A := C$, where $A$ is a primitive concept and $C$ is a concept. A concept definition allows stating the equivalence between concepts. For instance, Tennis:= SportKind $\sqcap$ ($\exists$HasSportTool.$\top$) $\sqcap$ ($\forall$HasSportTool.TennisRacket) states that tennis is identified by a kind of sport having a tennis racket as a sport tool. From a semantics point of view, an interpretation $\mathcal{I}$ *satisfies* a concept specialisation $A{<}C$ iff $A^{\mathcal{I}} \subseteq C^{\mathcal{I}}$. Similarly, an interpretation $\mathcal{I}$ *satisfies* a concept definition $A := C$ iff $A^{\mathcal{I}} = C^{\mathcal{I}}$.

### 2.5 Knowledge Base, Entailment and Subsumption

A finite set $K$ of assertions and terminological axioms will be called a *Knowledge Base* (KB). With $K_A$ we will denote the set of assertions in $K$, whereas with $K_T$ we will denote the set of terminological axioms in $K$, also called a *terminology*. A KB $K$ is *purely assertional* if $K_T = \emptyset$. Further, we will assume that a terminology $K_T$ is such that no concept $A$ appears more than once on the left hand side of a terminological axiom $\tau \in K_T$ and that no cyclic definitions are present in $K_T$.[3]

We will say that an interpretation $\mathcal{I}$ *satisfies* (*is a model of*) a KB $K$ iff $\mathcal{I}$ satisfies each element in $K$. A KB $K$ *entails* an assertion $\alpha$ (denoted by $K \models \alpha$) iff every model of $K$ also satisfies $\alpha$. Furthermore, let $K_T$ be a terminology and let $C, D$ be two concepts. We will say that $D$ *subsumes* $C$ with respect to (w.r.t.) $K_T$ (denoted by $C \sqsubseteq_{K_T} D$) iff for every model $\mathcal{I}$ of $K_T$, $C^{\mathcal{I}} \subseteq D^{\mathcal{I}}$ holds.

The problem of determining whether $K \models \alpha$ is called *entailment problem*; the problem of determining whether $C \sqsubseteq_{K_T} D$ is called *subsumption problem*; and the problem of determining whether $K$ is satisfiable is called *satisfiability problem*.

It is well known (Buchheit, Donini, & Schaerf, 1993a; Donini, Lenzerini, Nardi, & Schaerf, 1994; Nebel, 1990) that in $\mathcal{ALC}$

$$K \models (a,b){:}R \quad \text{iff} \quad (a,b){:}R \in K \tag{10}$$

$$K \models a{:}C \quad \text{iff} \quad K \cup \{a{:}\neg C\} \text{ is not satisfiable} \tag{11}$$

$$C \sqsubseteq_{\emptyset} D \quad \text{iff} \quad a{:}C \models a{:}D, \text{ for a new } a \tag{12}$$

$$C \sqsubseteq_{K_T} D \quad \text{iff} \quad C' \sqsubseteq_{\emptyset} D' \tag{13}$$

where $C'$ and $D'$ are build from $C$ and $D$ by expanding the terminology $K_T$ to $K_T''$ and substituting every primitive concept occurring in $C$ or $D$, which is defined in $K_T''$, with its defining term in $K_T''$. The *expansion of a KB K* works as follows (Nebel, 1990).

1. *Elimination of concept specialisation*: each concept specialisation $A{<}C \in K_T$ is replaced with a concept definition $A := C \sqcap A^*$, where $A^*$ is a new primitive concept. $A^*$ stands for the absent part of the definition of $A$. Let $K_T'$ be the terminology, which is obtained by replacing all concept specialisation by concept definitions.

---

3. We will say that $A$ *directly uses* primitive concept $B$ in $K_T$, if there is $\tau \in K_T$ such that $A$ is on the left hand side of $\tau$ and $B$ occurs in the right hand side of $\tau$. Let *uses* be the transitive closure of the relation directly uses in $K_T$. $K_T$ is *cyclic* iff there is $A$ such that $A$ uses $A$ in $K_T$.

2. *Expansion of $K'_T$*: every defined concept (*i.e.* the first argument of a concept definition) which occurs in the defining term of a concept definition (*i.e.* the second argument of a concept definition) is substituted by its defining term. This process is iterated until there remain only undefined concepts in the second arguments of concept definitions. This yields a terminology $K''_T$.

3. *Expansion of $K_A$*: every primitive concept occurring in $K_A$ which is defined in $K''_T$ is substituted by its defining term in $K''_T$. This yields $K'_A$.

The transformation has the nice property that $K \models \alpha$ iff $K'_A \models \alpha'$, where $\alpha'$ is obtained by replacing every primitive concept occurring in $\alpha$, which is defined in $K''_T$, with its defining term in $K''_T$. While this allows us to restrict our attention to purely assertional KBs only, it is worth noting that the expansion process can be exponential (Nebel, 1988).

From (10)–(13), it follows that the above problems can be reduced to the satisfiability problem. There exists a well known technique based on constraint propagation solving this problem (Schmidt-Schauß & Smolka, 1991; Buchheit, Donini, & Schaerf, 1993b; Donini et al., 1994).

We conclude with an example.

**Example 1** *Consider the following terminology $K_T$.*

| | | |
|---|---|---|
| SportKind | $<$ | $\top$ |
| SportTool | $<$ | $\top$ |
| IndividualSport | $<$ | SportKind |
| TeamSport | $<$ | SportKind |
| Basketball | $<$ | SportTool |
| TennisRacket | $<$ | SportTool |
| Basket | $:=$ | SportKind⊓ |
| | | ($\exists$KindOfSport.$\top$)⊓ |
| | | ($\forall$KindOfSport.TeamSport)⊓ |
| | | ($\exists$HasSportTool.$\top$)⊓ |
| | | ($\forall$HasSportTool.Basketball) |
| Tennis | $:=$ | SportKind⊓ |
| | | ($\exists$KindOfSport.$\top$)⊓ |
| | | ($\forall$KindOfSport.IndividualSport)⊓ |
| | | ($\exists$HasSportTool.$\top$)⊓ |
| | | ($\forall$HasSportTool.TennisRacket) |

*Suppose that there are two video sequences* v1*,* v2*, which are about basket and tennis, respectively. We may represent the semantic content of them through*

$$K_{v1} = \{v1{:}\mathsf{Video} \sqcap \exists\mathsf{About.Basket}\}$$
$$K_{v2} = \{v2{:}\mathsf{Video} \sqcap \exists\mathsf{About.Tennis}\}.$$

*Consider $K = K_T \cup K_{v1} \cup K_{v2}$. If we are interested in retrieving videos about sport, we may query $K$ through the query concept* $\mathsf{Q} = \mathsf{Video} \sqcap \exists\mathsf{About.SportKind}$ *and the answer will be the list containing both* v1 *and* v2*, as $K \models v1{:}\mathsf{Q}$ and $K \models v2{:}\mathsf{Q}$ hold.*

*On the other hand, if we are looking for individual sport videos, then, given the query concept* $\mathsf{Q'} = \mathsf{Video} \sqcap \exists\mathsf{About}.\exists\mathsf{KindOfSport}.\mathsf{IndividualSport}$, *it follows that only video* v2 *will be retrieved. In fact,* $K \not\models v1{:}\mathsf{Q'}$ *and* $K \models v2{:}\mathsf{Q'}$ *hold.* □

## 3. A Fuzzy DL

Our fuzzy extension directly relates to Zadeh's work on fuzzy sets (Zadeh, 1965). A *fuzzy set* $S$ with respect to an universe $U$ is characterised by a *membership function* $\mu_S : U \rightarrow [0,1]$, assigning an $S$-membership degree, $\mu_S(u)$, to each element $u$ in $U$. $\mu_S(u)$ gives us an estimation of the belonging of $u$ to $S$. Typically, if $\mu_S(u) = 1$ then $u$ definitely belongs to $S$, while $\mu_S(u) = 0.8$ means that $u$ is "likely" to be an element of $S$. Moreover, according to Zadeh, the membership function has to satisfy three well known restrictions: for all $u \in U$ and for all fuzzy sets $S_1, S_2$ with respect to $U$

$$\begin{aligned}
\mu_{S_1 \cap S_2}(u) &= \min\{\mu_{S_1}(u), \mu_{S_2}(u)\} \\
\mu_{S_1 \cup S_2}(u) &= \max\{\mu_{S_1}(u), \mu_{S_2}(u)\} \\
\mu_{\overline{S_1}}(u) &= 1 - \mu_{S_1}(u) \ ,
\end{aligned}$$

where $\overline{S_1}$ is the complement of $S_1$ in $U$. Alternative restrictions on membership functions have been proposed in the literature, but it is not our aim to investigate them here (the interested reader may consult *e.g.* Dubois & Prade, 1980).

A justification of the choice of the min and the max was given by Bellman and Giertz (1973), which have shown that under certain reasonable conditions min and max are the unique possible choice for set intersection and set union, respectively.

When we switch to fuzzy logics, the notion of degree of membership $\mu_S(u)$ of an element $u \in U$ w.r.t. the fuzzy set $S$ over $U$ is regarded as the *truth-value* of the statement *"u is S"*. Accordingly, in our fuzzy DL, (*i*) a concept $C$, rather than being interpreted as a classical set, will be interpreted as a fuzzy set and, thus, concepts become *imprecise*; and, consequently, (*ii*) the statement "a is C", *i.e.* a:C, will have a truth-value in $[0,1]$ given by the degree of membership of being the individual $a$ a member of the fuzzy set $C$.

### 3.1 Fuzzy Interpretation

A *fuzzy interpretation* is now a pair $\mathcal{I} = (\Delta^{\mathcal{I}}, \cdot^{\mathcal{I}})$, where $\Delta^{\mathcal{I}}$ is, as for the crisp case, the *domain*, whereas $\cdot^{\mathcal{I}}$ is an *interpretation function* mapping

1. individuals as for the crisp case, *i.e.* $a^{\mathcal{I}} \neq b^{\mathcal{I}}$, if $a \neq b$;

2. a concept $C$ into a membership function $C^{\mathcal{I}} : \Delta^{\mathcal{I}} \rightarrow [0,1]$;

3. a role $R$ into a membership function $R^{\mathcal{I}} : \Delta^{\mathcal{I}} \times \Delta^{\mathcal{I}} \rightarrow [0,1]$.

If $C$ is a concept then $C^{\mathcal{I}}$ will naturally be interpreted as the *membership degree function* of the fuzzy concept (set) $C$ w.r.t. $\mathcal{I}$, *i.e.* if $d \in \Delta^{\mathcal{I}}$ is an object of the domain $\Delta^{\mathcal{I}}$ then $C^{\mathcal{I}}(d)$ gives us the degree of being the object $d$ an element of the fuzzy concept $C$ under the interpretation $\mathcal{I}$. Similarly for roles. Additionally, the interpretation function $\cdot^{\mathcal{I}}$ has to satisfy the following equations: for all $d \in \Delta^{\mathcal{I}}$,

$$\begin{aligned}
\top^{\mathcal{I}}(d) &= 1 \\
\bot^{\mathcal{I}}(d) &= 0 \\
(C \sqcap D)^{\mathcal{I}}(d) &= \min\{C^{\mathcal{I}}(d), D^{\mathcal{I}}(d)\} \\
(C \sqcup D)^{\mathcal{I}}(d) &= \max\{C^{\mathcal{I}}(d), D^{\mathcal{I}}(d)\} \\
(\neg C)^{\mathcal{I}}(d) &= 1 - C^{\mathcal{I}}(d) \\
(\forall R.C)^{\mathcal{I}}(d) &= \inf_{d' \in \Delta^{\mathcal{I}}}\{\max\{1 - R^{\mathcal{I}}(d, d'), C^{\mathcal{I}}(d')\}\} \\
(\exists R.C)^{\mathcal{I}}(d) &= \sup_{d' \in \Delta^{\mathcal{I}}}\{\min\{R^{\mathcal{I}}(d, d'), C^{\mathcal{I}}(d')\}\}.
\end{aligned}$$

These equations are the standard interpretation of conjunction, disjunction, negation and quantification, respectively (see Lee, 1972; Tresp & Molitor, 1998).

Note that the semantics of $\exists R.C$

$$(\exists R.C)^{\mathcal{I}}(d) = \sup_{d' \in \Delta^{\mathcal{I}}}\{\min\{R^{\mathcal{I}}(d, d'), C^{\mathcal{I}}(d')\}\} \tag{14}$$

is the result of viewing $\exists R.C$ as the open first order formula $\exists y. F_R(x, y) \wedge F_C(y)$ (see (9)) and the existential quantifier $\exists$ is viewed as a disjunction over the elements of the domain. Similarly,

$$(\forall R.C)^{\mathcal{I}}(d) = \inf_{d' \in \Delta^{\mathcal{I}}}\{\max\{1 - R^{\mathcal{I}}(d, d'), C^{\mathcal{I}}(d')\}\} \tag{15}$$

is related to the open first order formula $\forall y. \neg F_R(x, y) \vee F_C(y)$ (see (8)), where the universal quantifier $\forall$ is viewed as a conjunction over the elements of the domain.

We will say that two concepts $C$ and $D$ are said to be *equivalent* (denoted by $C \cong D$) when $C^{\mathcal{I}} = D^{\mathcal{I}}$ for all interpretations $\mathcal{I}$. As for the crisp non fuzzy case, dual relationships between concepts hold: *e.g.* $\top \cong \neg \bot$, $(C \sqcap D) \cong \neg(\neg C \sqcup \neg D)$ and $(\forall R.C) \cong \neg(\exists R.\neg C)$.

## 3.2 Fuzzy Assertion

A *fuzzy assertion* (denoted by $\psi$) is an expression having one of the following form $\langle \alpha \geq n \rangle$ or $\langle \alpha \leq m \rangle$, where $\alpha$ is an $\mathcal{ALC}$ assertion, $n \in (0, 1]$ and $m \in [0, 1)$. From a semantics point of view, a fuzzy assertion $\langle \alpha \leq n \rangle$ constrains the truth-value of $\alpha$ to be less or equal to $n$ (similarly for $\geq$). Consequently, *e.g.* $\langle$v1:Video $\sqcap \exists$About.Basket $\geq 0.8\rangle$ states that video v1 is likely about basket. Formally, an interpretation $\mathcal{I}$ *satisfies* a fuzzy assertion $\langle a{:}C \geq n \rangle$ (resp. $\langle (a, b){:}R \geq n \rangle$) iff $C^{\mathcal{I}}(a^{\mathcal{I}}) \geq n$ (resp. $R^{\mathcal{I}}(a^{\mathcal{I}}, b^{\mathcal{I}}) \geq n$). Similarly, an interpretation $\mathcal{I}$ *satisfies* a fuzzy assertion $\langle a{:}C \leq n \rangle$ (resp. $\langle (a, b){:}R \leq n \rangle$) iff $C^{\mathcal{I}}(a^{\mathcal{I}}) \leq n$ (resp. $R^{\mathcal{I}}(a^{\mathcal{I}}, b^{\mathcal{I}}) \leq n$). Two fuzzy assertions $\psi_1$ and $\psi_2$ are said to be *equivalent* (denoted by $\psi_1 \cong \psi_2$) iff they are satisfied by the same set of interpretations. Notice that the combination of both $\langle a{:}C \geq m \rangle$ and $\langle a{:}C \leq n \rangle$, with $m \leq n$, restricts the truth-value of $a{:}C$ in between $[m, n]$. Moreover, $\langle a{:}\neg C \geq n \rangle \cong \langle a{:}C \leq 1 - n \rangle$. A *primitive fuzzy assertion* is a fuzzy assertion involving a primitive assertion.

One might wonder why we do not allow expressions of the form $\langle \alpha > n \rangle$ or the form $\langle \alpha < n \rangle$. The reason simply relies on the observation that it is quite hard to imagine situations in which we are able to assert such *strict* $>, <$ relations. So we will leave them out for ease.[4]

---

4. Of course, the whole can easily be extended in case we would like to consider these two types of assertions too.

Note that in the work of Straccia (1998), no fuzzy assertion of the form $\langle \alpha \leq n \rangle$ is allowed.

### 3.3 Fuzzy Terminological Axiom

Fuzzy terminological axioms we will consider are a natural extension of classical terminological axioms to the fuzzy case. From a syntax point of view, a *fuzzy terminological axiom* (denoted by $\tilde{\tau}$) is either a fuzzy concept specialisation or a fuzzy concept definition. A *fuzzy concept specialisation* is an expression of the form $A \prec C$, where $A$ is a primitive concept and $C$ is a concept. On the other hand, a *fuzzy concept definition* is an expression of the form $A :\approx C$, where $A$ is a primitive concept and $C$ is a concept. From a semantics point of view, we consider the natural extension of classical set inclusion to the fuzzy case (Zadeh, 1965). A fuzzy interpretation $\mathcal{I}$ *satisfies* a fuzzy concept specialisation $A \prec C$ iff

$$\forall d \in \Delta^{\mathcal{I}}, A^{\mathcal{I}}(d) \leq C^{\mathcal{I}}(d), \tag{16}$$

whereas $\mathcal{I}$ *satisfies* a fuzzy concept definition $A :\approx C$ iff

$$\forall d \in \Delta^{\mathcal{I}}, A^{\mathcal{I}}(d) = C^{\mathcal{I}}(d). \tag{17}$$

Note that in the work of Straccia (1998) a fuzzy specialisation is "non-standard". Indeed, Straccia (1998) considered fuzzy specialisations of the form $\langle A \to C \geq n \rangle$ where $(A \to C)^{\mathcal{I}} = \min_{d \in \Delta^{\mathcal{I}}} \{\max\{1 - A^{\mathcal{I}}(d), C^{\mathcal{I}}(d)\}\}$. A drawback of this formulation is that it is not clear where the $n$ in $\langle A \to C \geq n \rangle$ comes from, *i.e.* who defines the value $n$ and how it is determined. We prefer to rely here on the standard interpretation of fuzzy subsets.

### 3.4 Fuzzy Knowledge Base, Fuzzy Entailment and Fuzzy Subsumption

A *fuzzy knowledge base* is a finite set of fuzzy assertions and fuzzy terminological axioms. As for the crisp case, with $\Sigma_A$ we will denote the set of fuzzy assertions in $\Sigma$, with $\Sigma_T$ we will denote the set of fuzzy terminological axioms in $\Sigma$ (the terminology), if $\Sigma_T = \emptyset$ then $\Sigma$ is *purely assertional*, and we will assume that a terminology $\Sigma_T$ is such that no concept $A$ appears more than once on the left hand side of a fuzzy terminological axiom $\tilde{\tau} \in \Sigma_T$ and that no cyclic definitions are present in $\Sigma_T$.

An interpretation $\mathcal{I}$ *satisfies* (is *a model of*) a set of fuzzy $\Sigma$ iff $\mathcal{I}$ satisfies each element of $\Sigma$. A fuzzy KB $\Sigma$ *fuzzy entails* a fuzzy assertion $\psi$ (denoted by $\Sigma \models \psi$) iff every model of $\Sigma$ also satisfies $\psi$.

Furthermore, let $\Sigma_T$ be a terminology and let $C, D$ be two concepts. We will say that $D$ *fuzzy subsumes* $C$ w.r.t. $\Sigma_T$ (denoted by $C \preceq_{\Sigma_T} D$) iff for every model $\mathcal{I}$ of $\Sigma_T$, $\forall d \in \Delta^{\mathcal{I}}, C^{\mathcal{I}}(d) \leq D^{\mathcal{I}}(d)$ holds.

Finally, given a fuzzy KB $\Sigma$ and an assertion $\alpha$, it is of interest to compute $\alpha$'s best lower and upper truth-value bounds. To this end we define the *greatest lower bound* of $\alpha$ w.r.t. $\Sigma$ (denoted by $glb(\Sigma, \alpha)$) to be $\sup\{n : \Sigma \models \langle \alpha \geq n \rangle\}$. Similarly, we define the *least upper bound* of $\alpha$ with respect to $\Sigma$ (denoted by $lub(\Sigma, \alpha)$) to be $\inf\{n : \Sigma \models \langle \alpha \leq n \rangle\}$ ($\sup \emptyset = 0, \inf \emptyset = 1$). Determining the *lub* and the *glb* is called the *Best Truth-Value Bound* (BTVB) problem.

## 4. Some Properties

In this section we discuss some properties of our fuzzy logic. Several properties described by Straccia (2000b) for the propositional case are easily extended to our first order case too.

### 4.1 Concept Equivalence

The first ones are straightforward: $\neg\top \cong \bot$, $C \sqcap \top \cong C$, $C \sqcup \top \cong \top$, $C \sqcap \bot \cong \bot$, $C \sqcup \bot \cong C$, $\neg\neg C \cong C$, $\neg(C \sqcap D) \cong \neg C \sqcup \neg D$, $\neg(C \sqcup D) \cong \neg C \sqcap \neg D$, $C_1 \sqcap (C_2 \sqcup C_3) \cong (C_1 \sqcap C_2) \sqcup (C_1 \sqcap C_3)$ and $C_1 \sqcup (C_2 \sqcap C_3) \cong (C_1 \sqcup C_2) \sqcap (C_1 \sqcup C_3)$. For concepts involving roles, we have $\forall R.C \cong \neg\exists R.\neg C$, $\forall R.\top \cong \top$, $\exists R.\bot \cong \bot$ and $(\forall R.C) \sqcap (\forall R.D) \cong \forall R.(C \sqcap D)$. Please, note that we do not have $C \sqcap \neg C \cong \bot$, nor we have $C \sqcup \neg C \cong \top$ and, thus, $(\exists R.C) \sqcap (\forall R.\neg C) \not\cong \bot$ and $(\exists R.C) \sqcup (\forall R.\neg C) \not\cong \top$ hold. In general we can only say that $(C \sqcap \neg C)^{\mathcal{I}}(d) \leq 0.5$, for any interpretation $\mathcal{I}$ and $d \in \Delta^{\mathcal{I}}$ and, similarly, $(C \sqcup \neg C)^{\mathcal{I}}(d) \geq 0.5$, *i.e.* $lub(\emptyset, a{:}\neg C \sqcap C) = 0.5$ and $glb(\emptyset, a{:}\neg C \sqcup C) = 0.5$, respectively.

### 4.2 Entailment Relation

Of course, $\Sigma \models \langle \alpha \geq n \rangle$ iff $glb(\Sigma, \alpha) \geq n$, and similarly $\Sigma \models \langle \alpha \leq n \rangle$ iff $lub(\Sigma, \alpha) \leq n$ hold. Concerning roles, note that $\Sigma \models \langle (a,b){:}R \geq n \rangle$ iff $\langle (a,b){:}R \geq m \rangle \in \Sigma$ with $m \geq n$. Therefore,

$$glb(\Sigma, R(a,b)) = \max\{n : \langle R(a,b) \geq n \rangle \in \Sigma\} \tag{18}$$

while the same is not true for the $\leq$ case. While $\langle (a,b){:}R \leq m \rangle \in \Sigma$ and $m \leq n$ imply $\Sigma \models \langle (a,b){:}R \leq n \rangle$, the converse is false (*e.g.* $\{\langle a{:}\forall R.A \geq 1 \rangle, \langle b{:}A \leq 0 \rangle\} \models \langle (a,b){:}R \leq 0 \rangle$).

Furthermore, from $\Sigma \models \langle a{:}C \leq n \rangle$ iff $\Sigma \models \langle a{:}\neg C \geq 1 - n \rangle$,

$$
\begin{aligned}
1 - lub(\Sigma, C(a)) &= 1 - \inf\{n : \Sigma \models \langle a{:}C \leq n \rangle\} \\
&= \sup\{1 - n : \Sigma \models \langle a{:}C \leq n \rangle\} \\
&= \sup\{n : \Sigma \models \langle a{:}C \leq 1 - n \rangle\} \\
&= \sup\{n : \Sigma \models \langle a{:}\neg C \geq n \rangle\} \\
&= glb(\Sigma, a{:}\neg C).
\end{aligned}
$$

follows. Therefore,

$$lub(\Sigma, a{:}C) = 1 - glb(\Sigma, a{:}\neg C), \tag{19}$$

*i.e.* *lub* can be determined through *glb* (and vice-versa). The same reduction to *glb* does not hold for $lub(\Sigma, (a,b){:}R)$ as $(a,b){:}\neg R$ is not an expression of our language.[5]

*Modus ponens on concepts* is supported: if $m > 1 - n$ then $\{\langle a{:}C \geq m \rangle, \langle a{:}\neg C \sqcup D \geq n \rangle\} \models \langle a{:}D \geq n \rangle$ holds.

*Modus ponens on roles* is supported: if $m > 1 - n$ then $\{\langle (a,b){:}R \geq m \rangle, \langle a{:}\forall R.D \geq n \rangle\} \models \langle a{:}D \geq n \rangle$ and $\{\langle a{:}\exists R.C \geq m \rangle, \langle a{:}\forall R.D \geq n \rangle\} \models \langle a{:}\exists R.(C \sqcap D) \geq \min\{n,m\} \rangle$ hold. Moreover, $\{\langle a{:}\forall R.C \geq m \rangle, \langle a{:}\forall R.D \geq n \rangle\} \models \langle a{:}\forall R.(C \sqcap D) \geq \min\{n,m\} \rangle$ holds.

*Modus ponens on specialisation* is supported. The following degree bounds propagation through a taxonomy is supported. If $C \preceq_{\Sigma} D$ then (*i*) $\Sigma \cup \{\langle a{:}C \geq n \rangle\} \models \langle a{:}D \geq n \rangle$; and (*ii*) $\Sigma \cup \{\langle a{:}D \leq n \rangle\} \models \langle a{:}C \leq n \rangle$ hold.

---

5. Of course, $lub(\Sigma, (a,b){:}R) = 1 - glb(\Sigma, (a,b){:}\neg R)$ holds, where $(\neg R)^{\mathcal{I}}(d, d') = 1 - R^{\mathcal{I}}(d, d')$.

Note that, according to Straccia (1998)

$$\text{if } m > 1 - n \text{ then } \{\langle a{:}A \geq m \rangle, \langle A \to C \geq n \rangle\} \not\approx \langle a{:}C \geq n \rangle.$$

A drawback of the above property is that whatever the degree $m$ is (as long as $m > 1 - n$), from $\langle a{:}A \geq m \rangle$ and $\langle A \to C \geq n \rangle$ we infer $\langle a{:}C \geq n \rangle$, where $n$ is a priori fixed value.

### 4.3 Soundness of the Semantics

Our fuzzy semantics is *sound* w.r.t. crisp semantics. In fact, let $\Sigma$ be a fuzzy KB in which no $\langle (a, b){:}R \leq n \rangle$ occurs. We leave these fuzzy assertions out, as role negation is not present in crisp $\mathcal{ALC}$. Let us consider the following transformation $\sharp(\cdot)$ of fuzzy assertions into assertions, where $\sharp(\cdot)$ takes the "crisp" assertional part of a fuzzy assertion:

$$\sharp\langle \alpha \geq n \rangle \;\; \mapsto \;\; \alpha$$
$$\sharp\langle a{:}C \leq n \rangle \;\; \mapsto \;\; a{:}\neg C.$$

We extend $\sharp(\cdot)$ to fuzzy terminological axioms as follows: $\sharp\tilde{\tau} = \tau$. Finally, $\sharp\Sigma = \{\sharp\psi : \psi \in \Sigma\} \cup \{\sharp\tilde{\tau} : \tilde{\tau} \in \Sigma_T\}$.

It is quite easily to verify that

**Proposition 1** *Let $\Sigma$ be a fuzzy KB in which no $\langle (a, b){:}R \leq n \rangle$ occurs and let $\psi$ be a fuzzy assertion $\langle \alpha \geq n \rangle$. If $\Sigma \not\approx \langle \alpha \geq n \rangle$ then $\sharp\Sigma \models \sharp\psi$ (i.e. there cannot be fuzzy entailment without entailment in $\mathcal{ALC}$).* ⊣

*Proof:* Consider a classical interpretation $\mathcal{I}$ satisfying $\sharp\Sigma$. $\mathcal{I}$ is also a fuzzy interpretation such that $C^{\mathcal{I}}(d) \in \{0, 1\}$, $R^{\mathcal{I}}(d, d') \in \{0, 1\}$ hold. By induction on the structure of a concept $C$ it can be shown that $\mathcal{I}$ (classically) satisfies $a{:}C$ iff $C^{\mathcal{I}}(a^{\mathcal{I}}) = 1$. Similarly for roles. Therefore, $\mathcal{I}$ is a fuzzy interpretation satisfying $\Sigma$. By hypothesis, $\mathcal{I}$ satisfies $\langle \alpha \geq n \rangle$ and $n > 0$. Therefore, the truth-value of $\alpha$ under $\mathcal{I}$ is 1, *i.e.* $\mathcal{I}$ satisfies $\alpha$. □

For the general case, $\Sigma$ has to be satisfiable as $\langle (a, b){:}R \leq n \rangle$ may introduce an inconsistency, *e.g.* $\{\langle (a, b){:}R \leq 0.3 \rangle, \langle (a, b){:}R \geq 0.4 \rangle\} \not\approx \langle a{:}A \geq 1 \rangle$, but $\{(a, b){:}R\} \not\models a{:}A$.

The converse of Proposition 1 does not hold in general.

**Example 2** *Let $\Sigma$ be the set $\Sigma = \{\langle a{:}A \sqcup B \leq 0.6 \rangle, \langle a{:}A \geq 0.3 \rangle\}$. It follows that $\sharp\Sigma = \{a{:}\neg(A \sqcup B), a{:}A\}$ which is unsatisfiable. Therefore, it can easily be verified that $\sharp\Sigma \models a{:}B$, but $\Sigma \not\approx \langle a{:}B \geq n \rangle$, for all $n > 0$.* □

Once we restrict the attention to normalised fuzzy assertions, a converse relation follows immediately (Lee, 1972; Straccia, 2000b).

Indeed, we say that a fuzzy assertion $\psi$ is *KB-normalised* iff

1. if $\psi$ is $\langle \alpha \geq n \rangle$ then $n > 0.5$;

2. if $\psi$ is $\langle \alpha \leq n \rangle$ then $n < 0.5$.

We say that a fuzzy assertion $\psi$ is *query-normalised* iff

1. if $\psi$ is $\langle \alpha \geq n \rangle$ then $n \leq 0.5$;

2. if $\psi$ is $\langle \alpha \leq n \rangle$ then $n \geq 0.5$.

Note that the definitions for KB-normalisation and query-normalisation are dual. The following proposition follows from (Lee, 1972; Straccia, 2000b) and relies on the fact that $\langle \alpha \geq n \rangle$ and $\langle \alpha \leq m \rangle$ are together inconsistent, if $n > 0.5$ and $m < 0.5$. In particular, Straccia (2000b) shows that if $\Sigma$ and $\psi$ are a normalised fuzzy propositional KB and a query-normalised fuzzy proposition, then Then $\Sigma \not\approx \psi$ iff $\sharp\Sigma \models \sharp\psi$. The proof is given by showing that from a deduction proving $\Sigma \not\approx \psi$ a deduction proving $\sharp\Sigma \models \sharp\psi$ can be build and vice-versa. The extension to our case is straightforward as *e.g.* for $n, m > 0.5$, $\langle a{:}\forall R.C \geq n \rangle, \langle (a,b){:}R \geq m \rangle \not\approx \langle b{:}C \geq n \rangle$ iff $a{:}\forall R.C, (a,b){:}R \models b{:}C$ holds (the other first-order cases involving $\forall$ and $\exists$ are similar).

**Proposition 2** *Let $\Sigma$ be a fuzzy KB in which no $\langle (a,b){:}R \leq n \rangle$ occurs and each $\psi \in \Sigma$ is KB-normalised. Let $\psi$ be a query-normalised fuzzy assertion. Then $\Sigma \not\approx \psi$ iff $\sharp\Sigma \models \sharp\psi$.* ⊣

**Example 3** *Let $\Sigma$ be the set $\Sigma = \{\langle a{:}A \sqcap B \leq 0.4 \rangle, \langle a{:}A \geq 0.6 \rangle\}$. Let $\psi$ be $\langle a{:}B \leq 0.7 \rangle$. Each fuzzy assertion in $\Sigma$ is KB-normalised and $\psi$ is query-normalised. It follows that $\sharp\Sigma = \{a{:}\neg(A \sqcap B), a{:}A\}$ and $\sharp\psi = a{:}\neg B$. It is easily verified that $\Sigma \not\approx \psi$ and that $\sharp\Sigma \models \sharp\psi$, thereby confirming Proposition 2.* □

## 4.4 Subsumption

At first, as for the classical case and with the same method seen before, subsumption between two concepts $C$ and $D$ w.r.t. a terminology $\Sigma_T$, *i.e.* $C \preceq_{\Sigma_T} D$, can be reduced to the case of an empty terminology, *i.e.* $C' \preceq_\emptyset D'$.

**Example 4** *Suppose we have two images* i1 *and* i2 *each being a snapshot of the car traffic on a major street of an European city. An underlying image analysis tool recognizes, among all the recognised objects, that in image* i1 *there is a Ferrari, while in image* i2 *there is a Porsche. Furthermore, a semantic image indexing tool establishes that, to some degree n image* i1 *is about a Ferrari, whereas to some degree m image* i2 *is about a Porsche. Please note that, as a weight of a keyword in text is a quantitative description of the aboutness of the text w.r.t. the keyword, a truth-degree gives a quantitative description of the aboutness of an images w.r.t. an object,* i.e. *the aboutness is handled as an imprecise concept. So, let us consider*

$$\Sigma = \{\langle \mathsf{i1}{:}\exists \mathsf{About.Ferrari} \geq 0.6 \rangle, \langle \mathsf{i2}{:}\exists \mathsf{About.Porsche} \geq 0.8 \rangle,$$
$$\mathsf{Ferrari} \prec \mathsf{Car}, \mathsf{Porsche} \prec \mathsf{Car}\}.$$

*where the axioms specify that both a Ferrari and a Porsche are a car. According to the expansion process, $\Sigma$ will be replaced by*

$$\Sigma' = \{\langle \mathsf{i1}{:}\exists \mathsf{About.Ferrari} \geq 0.6 \rangle, \langle \mathsf{i2}{:}\exists \mathsf{About.Porsche} \geq 0.8 \rangle,$$
$$\mathsf{Ferrari}{:}\approx \mathsf{Car} \sqcap \mathsf{Ferrari}^*, \mathsf{Porsche}{:}\approx \mathsf{Car} \sqcap \mathsf{Porsche}^*\},$$

*which will be simplified to*

$$\Sigma'' = \{\langle \mathsf{i1{:}\exists About.(Car \sqcap Ferrari^*)} \geq 0.6\rangle,$$
$$\langle \mathsf{i2{:}\exists About.(Car \sqcap Porsche^*)} \geq 0.8\rangle\}.$$

*Now, if we are looking for images which are about cars, then from $\Sigma$ we may infer that $\Sigma \vDash \langle \mathsf{i1{:}\exists About.Car} \geq 0.6\rangle$ and $\Sigma \vDash \langle \mathsf{i2{:}\exists About.Car} \geq 0.8\rangle$. Furthermore, it is easily verified that $\Sigma'' \vDash \langle \mathsf{i1{:}\exists About.Car} \geq 0.6\rangle$ and $\Sigma'' \vDash \langle \mathsf{i2{:}\exists About.Car} \geq 0.8\rangle$ hold as well. Indeed, for any fuzzy assertion $\psi$, $\Sigma \vDash \psi$ iff $\Sigma'' \vDash \psi$ holds.* □

We conclude this section with the analogue of Example 1 for the fuzzy case..

**Example 5** Consider the terminology $K_T$ and the query concept $\mathsf{Q}$ in Example 1. Let us define $\Sigma_T$ as the fuzzy KB derived from $K_T$ in which each terminological axiom $\tau$ has been replaced with the fuzzy terminological axiom $\tilde{\tau}$. Moreover, let us suppose that an underlying semantic video indexing tool furnishes the following semantic descriptions of the two videos $\mathsf{v1}$ and $\mathsf{v2}$.

$$\Sigma_{\mathsf{v1}} = \{\langle \mathsf{v1{:}Video} \geq 1\rangle, \langle \mathsf{v1{:}\exists About.Basket} \geq 0.9\rangle\}$$
$$\Sigma_{\mathsf{v2}} = \{\langle \mathsf{v2{:}Video} \geq 1\rangle, \langle \mathsf{v2{:}\exists About.Tennis} \geq 0.6\rangle\},$$

*i.e.* video $\mathsf{v1}$ is about basket with degree 0.9, whereas video $\mathsf{v2}$ is about tennis with degree 0.6. Let us consider $\Sigma = \Sigma_T \cup \Sigma_{\mathsf{v1}} \cup \Sigma_{\mathsf{v2}}$. It is easily verified that $glb(\Sigma, \mathsf{v1{:}Q}) = 0.9$, whereas $glb(\Sigma, \mathsf{v2{:}Q}) = 0.6$ hold. Therefore, video $\mathsf{v1}$ will be ranked before video $\mathsf{v2}$ after the retrieval process. □

## 5. Decision Algorithms in Fuzzy $\mathcal{ALC}$

Deciding whether $\Sigma \vDash \langle \alpha \geq n\rangle$ or $\Sigma \vDash \langle \alpha \leq m\rangle$ requires a calculus. Without loss of generality we will consider purely assertional fuzzy KBs only.

We will develop a calculus in the style of the constraint propagation method, as this method is usually proposed in the context of DLs (Buchheit et al., 1993a). The calculus extends the fuzzy propositional calculus described by Chen and Kundu (1996) and by Straccia (2000b) to our fuzzy DL case. We first address the entailment problem, then the subsumption problem and finally the BTVB problem. Both the subsumption problem and the BTVB problem will be reduced to the entailment problem.

### 5.1 A Decision Procedure for the Entailment Problem

Consider a new alphabet of $\mathcal{ALC}$ *variables*. An interpretation is extended to variables by mapping these into elements of the interpretation domain. An $\mathcal{ALC}$ *object* (denoted by $w$) is either an individual or a variable.[6]

A *constraint* (denoted by $\alpha$) is an expression of the form $w{:}C$ or $(w, w'){:}R$, where $w, w'$ are objects, $C$ is an $\mathcal{ALC}$ concept and $R$ is a role. A *fuzzy constraint* (denoted by $\psi$) is an expression having one of the following four forms: $\langle \alpha \geq n\rangle, \langle \alpha > n\rangle, \langle \alpha \leq n\rangle, \langle \alpha < n\rangle$. Note that assertions and fuzzy assertions are constraints and fuzzy constraints, respectively.

---

6. In the following, if there is no ambiguity, $\mathcal{ALC}$ variables and $\mathcal{ALC}$ objects are called variables and objects, respectively.

The definitions of satisfiability of a constraint, a fuzzy constraint, a set of constraints, a set of fuzzy constraints, primitive constraint and primitive fuzzy constraint are obvious.

It is quite easily verified that the fuzzy entailment problem can be reduced to the unsatisfiability problem of a set of fuzzy constraints:

$$\Sigma \not\approx \langle \alpha \geq n \rangle \quad \text{iff} \quad \Sigma \cup \{\langle \alpha < n \rangle\} \text{ not satisfiable} \tag{20}$$

$$\Sigma \not\approx \langle \alpha \leq n \rangle \quad \text{iff} \quad \Sigma \cup \{\langle \alpha > n \rangle\} \text{ not satisfiable.} \tag{21}$$

Our calculus, determining whether a finite set $S$ of fuzzy constraints is satisfiable or not, is based on a set of constraint propagation rules transforming a set $S$ of fuzzy constraints into "simpler" satisfiability preserving sets $S_i$ until either all $S_i$ contain a clash (indicating that from all the $S_i$ no model of $S$ can be build) or some $S_i$ is completed and clash-free, that is, no rule can be further be applied to $S_i$ and $S_i$ contains no clash (indicating that from $S_i$ a model of $S$ can be build).

A set of fuzzy constraints $S$ contains a *clash* iff it contains either one of the constraints in Table 1 or $S$ contains a conjugated pair of fuzzy constraints. Each entry in Table 2 says us

$\langle w{:}\bot \geq n \rangle$, where $n > 0$
$\langle w{:}\top \leq n \rangle$, where $n < 1$
$\langle w{:}\bot > n \rangle$, $\langle w{:}\top < n \rangle$, $\langle w{:}C < 0 \rangle$, $\langle w{:}C > 1 \rangle$

Table 1: Clashes

|  | $\langle \alpha < m \rangle$ | $\langle \alpha \leq m \rangle$ |
|---|---|---|
| $\langle \alpha \geq n \rangle$ | $n \geq m$ | $n > m$ |
| $\langle \alpha > n \rangle$ | $n \geq m$ | $n \geq m$ |

Table 2: Conjugated Pairs

under which condition the row-column pair of fuzzy constraints is a *conjugated pair*. Given a fuzzy constraint $\psi$, with $\psi^c$ we indicate a conjugate of $\psi$ (if there exists one). Notice that a conjugate of a fuzzy constraint may be not unique, as there could be infinitely many. For instance, both $\langle a{:}C < 0.6 \rangle$ and $\langle a{:}C \leq 0.7 \rangle$ are conjugates of $\langle a{:}C \geq 0.8 \rangle$.

Concerning the rules, for each connective $\sqcap, \sqcup, \neg, \forall, \exists$ there is a rule for each relation $\geq, >, \leq, <$, *i.e.* there are 20 rules. The rules have the form:

$$\Phi \rightarrow \Psi \text{ if } \Gamma \tag{22}$$

where $\Phi$ and $\Psi$ are sequences of fuzzy constraints and $\Gamma$ is a condition. A rule fires only if *the condition $\Gamma$ holds, if the current set $S$ of fuzzy constraints contains fuzzy constraints matching the precondition $\Phi$ and the consequence $\Psi$ is not already in $S$.* After firing, the constraints from $\Psi$ are added to $S$. The rules are the following:

$$(\neg_{\geq}) \quad \langle w{:}\neg C \geq n \rangle \to \langle w{:}C \leq 1-n \rangle$$
$$(\neg_{>}) \quad \langle w{:}\neg C > n \rangle \to \langle w{:}C < 1-n \rangle$$
$$(\neg_{\leq}) \quad \langle w{:}\neg C \leq n \rangle \to \langle w{:}C \geq 1-n \rangle$$
$$(\neg_{<}) \quad \langle w{:}\neg C < n \rangle \to \langle w{:}C > 1-n \rangle$$

(23)

$$(\sqcap_{\geq}) \quad \langle w{:}C \sqcap D \geq n \rangle \to \langle w{:}C \geq n \rangle, \langle w{:}D \geq n \rangle$$
$$(\sqcap_{>}) \quad \langle w{:}C \sqcap D > n \rangle \to \langle w{:}C > n \rangle, \langle w{:}D > n \rangle$$
$$(\sqcup_{\leq}) \quad \langle w{:}C \sqcup D \leq n \rangle \to \langle w{:}C \leq n \rangle, \langle w{:}D \leq n \rangle$$
$$(\sqcup_{<}) \quad \langle w{:}C \sqcup D < n \rangle \to \langle w{:}C < n \rangle, \langle w{:}D < n \rangle$$

$$(\sqcup_{\geq}) \quad \langle w{:}C \sqcup D \geq n \rangle \to \langle w{:}C \geq n \rangle \mid \langle w{:}D \geq n \rangle$$
$$(\sqcup_{>}) \quad \langle w{:}C \sqcup D > n \rangle \to \langle w{:}C > n \rangle \mid \langle w{:}D > n \rangle$$
$$(\sqcap_{\leq}) \quad \langle w{:}C \sqcap D \leq n \rangle \to \langle w{:}C \leq n \rangle \mid \langle w{:}D \leq n \rangle$$
$$(\sqcap_{<}) \quad \langle w{:}C \sqcap D < n \rangle \to \langle w{:}C < n \rangle \mid \langle w{:}D < n \rangle$$

$$(\forall_{\geq}) \quad \langle w_1{:}\forall R.C \geq n \rangle, \psi^c \to \langle w_2{:}C \geq n \rangle$$
$$\text{if } \psi \text{ is } \langle (w_1, w_2){:}R \leq 1-n \rangle$$
$$(\forall_{>}) \quad \langle w_1{:}\forall R.C > n \rangle, \psi^c \to \langle w_2{:}C > n \rangle$$
$$\text{if } \psi \text{ is } \langle (w_1, w_2){:}R < 1-n \rangle$$
$$(\exists_{\leq}) \quad \langle w_1{:}\exists R.C \leq n \rangle, \psi^c \to \langle w_2{:}C \leq n \rangle$$
$$\text{if } \psi \text{ is } \langle (w_1, w_2){:}R \leq n \rangle$$
$$(\exists_{<}) \quad \langle w_1{:}\exists R.C < n \rangle, \psi^c \to \langle w_2{:}C < n \rangle$$
$$\text{if } \psi \text{ is } \langle (w_1, w_2){:}R < n \rangle$$

$(\exists_{\geq}) \quad \langle w{:}\exists R.C \geq n \rangle \to \langle (w,x){:}R \geq n \rangle, \langle x{:}C \geq n \rangle$

if $x$ new variable and there is no $w'$ such that both

$\langle (w,w'){:}R \geq n \rangle$ and $\langle w'{:}C \geq n \rangle$ are already in the constraint set

$(\exists_{>}) \quad \langle w{:}\exists R.C > n \rangle \to \langle (w,x){:}R > n \rangle, \langle x{:}C > n \rangle$

if $x$ new variable and there is no $w'$ such that both

$\langle (w,w'){:}R > n \rangle$ and $\langle w'{:}C > n \rangle$ are already in the constraint set

$(\forall_{\leq}) \quad \langle w{:}\forall R.C \leq n \rangle \to \langle (w,x){:}R \geq 1-n \rangle, \langle x{:}C \leq n \rangle$

if $x$ new variable and there is no $w'$ such that both

$\langle (w,w'){:}R \geq 1-n \rangle$ and $\langle w'{:}C \leq n \rangle$ are already in the constraint set

$(\forall_{<}) \quad \langle w{:}\forall R.C < n \rangle \to \langle (w,x){:}R > 1-n \rangle, \langle x{:}C < n \rangle$

if $x$ new variable and there is no $w'$ such that both

$\langle (w,w'){:}R > 1-n \rangle$ and $\langle w'{:}C < n \rangle$ are already in the constraint set

Examples of rule instances are the following:

$$(\forall_{\geq}) \quad \langle a{:}\forall R.C \geq 0.7\rangle, \langle (a,b){:}R \geq 0.6\rangle \Rightarrow \langle b{:}C \geq 0.7\rangle$$
$$\psi \text{ is } \langle (a,b){:}R \leq 0.3\rangle$$
$$\psi^c = \langle (a,b){:}R \geq 0.6\rangle \text{ is a conjugate of } \psi$$

$$(\exists_{<}) \quad \langle a{:}\exists R.C < 0.8\rangle, \langle (a,b){:}R \geq 0.9\rangle \Rightarrow \langle b{:}C < 0.8\rangle$$
$$\psi \text{ is } \langle (a,b){:}R < 0.8\rangle$$
$$\psi^c = \langle (a,b){:}R \geq 0.9\rangle \text{ is a conjugate of } \psi$$

$$(\exists_{\geq}) \quad \langle a{:}\exists R.C \geq 0.8\rangle \Rightarrow \langle (w,x){:}R \geq 0.8\rangle, \langle x{:}C \geq 0.8\rangle$$
$$x \text{ new variable}$$

$$(\forall_{<}) \quad \langle a{:}\forall R.C < 0.8\rangle \Rightarrow \langle (w,x){:}R > 0.2\rangle, \langle x{:}C < 0.8\rangle$$
$$x \text{ new variable.}$$

A set of fuzzy constraints $S$ is said to be *complete* if no rule is applicable to it. Any complete set of fuzzy constraints $S_2$ obtained from a set of fuzzy constraints $S_1$ by applying the above rules (23) is called a *completion* of $S_1$. Due to the rules $(\sqcup_{\geq})$, $(\sqcup_{>})$, $(\sqcap_{\leq})$ and $(\sqcap_{<})$, more than one completion can be obtained. These rules are called *nondeterministic rules*. All other rules are called *deterministic rules*.

It is easily verified that the above calculus has the *termination property*, *i.e.* any completion of a finite set of fuzzy constraints $S$ can be obtained after a finite number of rule applications.

**Example 6** Let us consider the following fuzzy KB:

$$\Sigma = \{\langle a{:}\exists R.D \geq 0.7\rangle, \langle a{:}\forall R.C \geq 0.4\rangle, \langle (a,b){:}R \geq 0.5\rangle, \langle b{:}C \geq 0.2\rangle, \langle b{:}D \geq 0.3\rangle\}$$

Let $\alpha$ be the assertion $a{:}\exists R.(D \sqcap C)$, let $\psi$ be the fuzzy assertion $\langle \alpha \geq 0.4\rangle$, whereas let $\psi'$ be the fuzzy assertion $\langle \alpha \geq 0.5\rangle$. It is easily verified that $\Sigma \not\approx \psi$, whereas $\Sigma \not\approx \psi'$. We show that $\Sigma \not\approx \psi'$, by verifying that there is a clash-free completion of $S = \Sigma \cup \{\langle a{:}\exists R.(D \sqcap C) < 0.5\rangle\}$ (precisely, there are two of them).

By applying rules (23), we have the following sequences.

| | | |
|---|---|---|
| (1) | $\langle a{:}\exists R.D \geq 0.7\rangle$ | Hypothesis:$S$ |
| (2) | $\langle a{:}\forall R.C \geq 0.4\rangle$ | |
| (3) | $\langle (a,b){:}R \geq 0.5\rangle$ | |
| (4) | $\langle b{:}C \geq 0.2\rangle$ | |
| (5) | $\langle b{:}D \geq 0.3\rangle$ | |
| (6) | $\langle a{:}\exists R.(D \sqcap C) < 0.5\rangle$ | |
| (7) | $\langle (a,x){:}R \geq 0.7\rangle, \langle x{:}D \geq 0.7\rangle$ | $(\exists_{\geq}) : (1)$ |
| (8) | $\langle x{:}C \geq 0.4\rangle$ | $(\forall_{\geq}) : (2),(7)$ |
| (9) | $\langle b{:}D \sqcap C < 0.5\rangle$ | $(\exists_{<}) : (3),(6)$ |
| (10) | $\langle x{:}D \sqcap C < 0.5\rangle$ | $(\exists_{<}) : (6),(7)$ |
| | $\Omega_1 \quad \mid \quad \Omega_2$ | |

where the two sequences $\Omega_1$ and $\Omega_2$ are defined as follows: for $\Omega_1$ we have the two sequences

$$(11) \quad \langle b{:}D < 0.5 \rangle \quad (\sqcap_<) : (9)$$

| (12) | $\langle x{:}D < 0.5 \rangle$ | $(\sqcap_<) : (10)$ | (14) | $\langle x{:}C < 0.5 \rangle$ | $(\sqcap_<) : (10)$ |
|---|---|---|---|---|---|
| (13) | clash | $(7),(12)$ | (15) | clash-free | |

and for $\Omega_2$ we have the two sequences

$$(16) \quad \langle b{:}C < 0.5 \rangle \quad (\sqcap_<) : (9)$$

| (17) | $\langle x{:}D < 0.5 \rangle$ | $(\sqcap_<) : (10)$ | (19) | $\langle x{:}C < 0.5 \rangle$ | $(\sqcap_<) : (10)$ |
|---|---|---|---|---|---|
| (18) | clash | $(7),(17)$ | (20) | clash-free | |

□

**Example 7** *Consider Example 4 and let us prove that $\Sigma'' \not\approx \langle (\exists\mathsf{About}.\mathsf{Car})(\mathsf{i1}) \geq 0.6 \rangle$. We prove the above relation by verifying that all completions of $S = \Sigma'' \cup \{\langle \mathsf{i1}{:}\exists\mathsf{About}.\mathsf{Car} < 0.6 \rangle\}$ contain a clash. In fact, we have the following sequence.*

| (1) | $\langle \mathsf{i1}{:}\exists\mathsf{About}.(\mathsf{Car} \sqcap \mathsf{Ferrari}^*) \geq 0.6 \rangle$ | *Hypothesis:S* |
|---|---|---|
| (2) | $\langle \mathsf{i2}{:}\exists\mathsf{About}.(\mathsf{Car} \sqcap \mathsf{Porsche}^*) \geq 0.8 \rangle$ | |
| (3) | $\langle \mathsf{i1}{:}\exists\mathsf{About}.\mathsf{Car} < 0.6 \rangle$ | |
| (4) | $\langle (\mathsf{i1},\mathsf{x}){:}\mathsf{About} \geq 0.6 \rangle, \langle \mathsf{x}{:}(\mathsf{Car} \sqcap \mathsf{Ferrari}^*) \geq 0.6 \rangle$ | $(\exists_\geq) : (1)$ |
| (5) | $\langle \mathsf{x}{:}\mathsf{Car} < 0.6 \rangle$ | $(\exists_<) : (3),(4)$ |
| (6) | $\langle \mathsf{x}{:}\mathsf{Car} \geq 0.6 \rangle, \langle \mathsf{x}{:}\mathsf{Ferrari}^* \geq 0.6 \rangle,$ | $(\sqcap) : (4)$ |
| (7) | *clash* | $(5),(6)$ |

□

**Proposition 3** *A finite set of fuzzy constraints $S$ is satisfiable iff there exists a clash free completion of $S$.* ⊣

*Proof:*

⇒ .) Given the termination property, it is easily verified, by case analysis, that the above rules are sound, *i.e.* if $S_1$ is satisfiable then there is a satisfiable completion $S_2$ of $S_1$ and, thus, $S_2$ contains no clash. For instance, let us show that the $(\forall_\geq)$ rule is sound. Assume $\mathcal{I}$ is an interpretation satisfying $\langle w_1{:}\forall R.C \geq n \rangle$ and $\langle (w_1,w_2){:}R \geq m \rangle$, where $m > 1 - n$. Let us show that $\mathcal{I}$ satisfies $\langle w_2{:}C \geq n \rangle$. Since $\mathcal{I}$ satisfies $\langle w_1{:}\forall R.C \geq n \rangle$ it follows that $\max\{1 - R^{\mathcal{I}}(w_1^{\mathcal{I}}, w_2^{\mathcal{I}}), C^{\mathcal{I}}(w_2^{\mathcal{I}})\} \geq n$. But, $R^{\mathcal{I}}(w_1^{\mathcal{I}}, w_2^{\mathcal{I}}) \geq m$ and, thus, $1 - R^{\mathcal{I}}(w_1^{\mathcal{I}}, w_2^{\mathcal{I}}) \leq 1 - m < n$. As a consequence, $C^{\mathcal{I}}(w_2^{\mathcal{I}}) \geq n$ follows, *i.e.* $\mathcal{I}$ satisfies $\langle w_2{:}C \geq n \rangle$.

⇐ .) Suppose that there exists a clash free completion $S'$ of $S$. We build from $S'$ an interpretation $\mathcal{I}$ satisfying $S'$ and, as $S \subseteq S'$, $\mathcal{I}$ satisfies $S$. $\mathcal{I}$ is called *canonical model*.

For any primitive constraint $\alpha \in S'$, we collect its lower and upper bound restrictions in $S'$ as follows: let

$$
\begin{aligned}
N^{\geq}[\alpha] &= \{n : \langle \alpha \geq n \rangle \in S'\} \\
N^{>}[\alpha] &= \{n : \langle \alpha > n \rangle \in S'\} \\
N^{\leq}[\alpha] &= \{n : \langle \alpha \leq n \rangle \in S'\} \\
N^{<}[\alpha] &= \{n : \langle \alpha < n \rangle \in S'\}.
\end{aligned}
$$

We have to define $\mathcal{I}$ such that for each constraint $\alpha$, $\mathcal{I}$ satisfies the constraints collected in the sets $N^{(\cdot)}[\alpha]$: given $N^{\geq}[\alpha]$, the truth value $n$ of $\alpha$ under $\mathcal{I}$ has to be such that $n \geq \max N^{\geq}[\alpha]$, whereas w.r.t. $N^{>}[\alpha]$, the truth value $n$ of $\alpha$ under $\mathcal{I}$ has to be such that $n \geq \max N^{>}[\alpha] + \epsilon$, for a $\epsilon > 0$. Similarly, for the other cases, for instance, w.r.t. $N^{<}[\alpha]$, the truth value $n$ of $\alpha$ under $\mathcal{I}$ has to be such that $n \leq \max N^{>}[\alpha] - \epsilon$, for a $\epsilon > 0$. The two tables below

| $N^{\geq}[\alpha]$ | $N^{>}[\alpha]$ | $glb[\alpha, \epsilon]$ |
|---|---|---|
| $\emptyset$ | $\emptyset$ | 0 |
| $\emptyset$ | $\neq \emptyset$ | $n' + \epsilon$ |
| $\neq \emptyset$ | $\emptyset$ | $n$ |
| $\neq \emptyset$ | $\neq \emptyset$ | if $n > n'$ then $n$ else $n' + \epsilon$ |

| $N^{\leq}[\alpha]$ | $N^{<}[\alpha]$ | $lub[\alpha, \epsilon]$ |
|---|---|---|
| $\emptyset$ | $\emptyset$ | 1 |
| $\emptyset$ | $\neq \emptyset$ | $m' - \epsilon$ |
| $\neq \emptyset$ | $\emptyset$ | $m$ |
| $\neq \emptyset$ | $\neq \emptyset$ | if $m < m'$ then $m$ else $m' - \epsilon$ |

define for any $\alpha \in S'$ and $\epsilon > 0$, $lub[\alpha, \epsilon]$ and $glb[\alpha, \epsilon]$, the lower and upper bound constraints which $\mathcal{I}$ has to satisfy. In the tables, with $n, n', m, m'$ we indicate $\max N^{\geq}[\alpha]$, $\max N^{>}[\alpha]$, $\min N^{\leq}[\alpha]$ and $\min N^{<}[\alpha]$, respectively. In each table we distinguish between the four cases where the sets are empty (no constraints) or not. For instance, if for a constraint $w{:}A$, only $\langle w{:}A \geq 0.3 \rangle$, $\langle w{:}A > 0.4 \rangle$, $\langle w{:}A \leq 0.5 \rangle$ and $\langle w{:}A < 0.6 \rangle$ are in $S'$, then according to the first table bellow (row 4), for a $\epsilon > 0$, $glb[w{:}A, \epsilon] = 0.4 + \epsilon$, whereas $lub[w{:}A, \epsilon] = 0.5$.

We will define $\mathcal{I}$ such that $A^{\mathcal{I}}(w^{\mathcal{I}}) = glb[w{:}A, \epsilon]$. To make sure that $glb[w{:}A, \epsilon] \leq A^{\mathcal{I}}(w^{\mathcal{I}}) \leq lub[w{:}A, \epsilon]$, we have to choose an $\epsilon > 0$ small enough such that $glb[w{:}A, \epsilon] \leq lub[w{:}A, \epsilon]$, *i.e.* $0.4 + \epsilon \leq 0.5$. The existence of such an $\epsilon > 0$ is guaranteed by the fact that $S'$ is clash-free. An additional condition that the choice of such an $\epsilon$ has to satisfy concerns the case of a constraint $\alpha$ of type $(w, w'){:}R$. Let us show the problem with an example. Suppose $S'$ is $\{\langle (w, w'){:}R > 0.3 \rangle, \langle w{:}\forall R.B \geq 0.6 \rangle, \langle w'{:}B \leq 0.5 \rangle\}$. Therefore, according to the above tables, $glb[(w, w'){:}R, \epsilon_1] = 0.3 + \epsilon_1$, $lub[(w, w'){:}R, \epsilon_1] = 1$, $glb[w'{:}B, \epsilon_2] = 0$ and $lub[w'{:}B, \epsilon_2] = 0.5$. So, it seems that it is sufficient to choose an $\epsilon_1 > 0$ such that $0.3 + \epsilon_1 \leq 1$, which is indeed not the case. In fact, $\langle w{:}\forall R.B \geq 0.6 \rangle$ and $\langle w'{:}B \leq 0.5 \rangle$ introduces an upper bound on $(w, w'){:}R$, *i.e.* the truth-value of $(w, w'){:}R$ under $\mathcal{I}$ has to be less or equal to $0.4 = 1 - 0.6$. That is, we have to choose an $\epsilon_1 > 0$ such that $0.3 + \epsilon_1 \leq 0.4$. Otherwise, the truth-value of $w'{:}B$ under $\mathcal{I}$ has to be greater or equal to 0.6, contradicting $lub[w'{:}B, \epsilon_2] = 0.5$. Again, the existence of such an $\epsilon$ is guaranteed as $S'$ is clash-free.

Summing up: since $S'$ is clash-free, it follows that for each primitive constraint $\alpha$, there is $\epsilon[\alpha] > 0$ such that

$$glb[\alpha, \epsilon[\alpha]] \leq lub[\alpha, \epsilon[\alpha]]$$

where if $\alpha$ is $(w, w'){:}R$ then

for each $\langle w{:}\forall R.C \geq n \rangle$, if $\langle w'{:}C \geq n \rangle \notin S'$ then $glb[\alpha, \epsilon[\alpha]] \leq 1 - n$;
for each $\langle w{:}\forall R.C > n \rangle$, if $\langle w'{:}C > n \rangle \notin S'$ then $glb[\alpha, \epsilon[\alpha]] < 1 - n$;
for each $\langle w{:}\exists R.C \leq n \rangle$, if $\langle w'{:}C \leq n \rangle \notin S'$ then $glb[\alpha, \epsilon[\alpha]] \leq n$;
for each $\langle w{:}\exists R.C < n \rangle$, if $\langle w'{:}C < n \rangle \notin S'$ then $glb[\alpha, \epsilon[\alpha]] < n$. $\qquad(24)$

Now, consider the following interpretation $\mathcal{I}$ such that

1. the domain $\Delta^{\mathcal{I}}$ is the set of objects appearing in $S'$;

2. $w^{\mathcal{I}} = w$, for all $w \in \Delta^{\mathcal{I}}$;

3. $\top^{\mathcal{I}}(w^{\mathcal{I}}) = 1$ and $\bot^{\mathcal{I}}(w^{\mathcal{I}}) = 0$, for all $w \in \Delta^{\mathcal{I}}$;

4. $A^{\mathcal{I}}(w^{\mathcal{I}}) = glb[w{:}A, \epsilon[w{:}A]]$, for all primitive concepts $A$ and for all $w \in \Delta^{\mathcal{I}}$; and

5. $R^{\mathcal{I}}(w^{\mathcal{I}}, w'^{\mathcal{I}}) = glb[(w, w'){:}R, \epsilon[(w, w'){:}R]]$, for all roles $R$ and for all $w, w' \in \Delta^{\mathcal{I}}$.

We show, on induction on the structure of fuzzy constraints $\psi \in S'$, that $\mathcal{I}$ satisfies $S'$.

**Case** $\langle w{:}A > n \rangle$**:** By definition, $A^{\mathcal{I}}(w^{\mathcal{I}}) = glb[w{:}A, \epsilon[w{:}A]] > n$ and, thus, $\mathcal{I}$ satisfies $\langle w{:}A > n \rangle$. The cases $\geq, \leq$ and $<$ are similar.

**Case** $\langle (w, w'){:}R \geq n \rangle$**:** By definition, $R^{\mathcal{I}}(w^{\mathcal{I}}, w'^{\mathcal{I}}) = glb[(w, w'){:}R, \epsilon[(w, w'){:}R]] \geq n$ and, thus, $\mathcal{I}$ satisfies $\langle (w, w'){:}R \geq n \rangle$. The cases $>, \leq$ and $<$ are similar.

**Case** $\langle w{:}C \sqcap D \geq n \rangle$**:** From $\langle w{:}C \sqcap D \geq n \rangle \in S'$ and $S'$ completed, $\langle w{:}C \geq n \rangle \in S'$ and $\langle w{:}D \geq n \rangle \in S'$ follows. By induction, $\mathcal{I}$ satisfies both $\langle w{:}C \geq n \rangle$ and $\langle w{:}D \geq n \rangle$ and, thus, $\mathcal{I}$ satisfies $\langle w{:}C \sqcap D \geq n \rangle$. The cases $>, \leq$ and $<$ are similar.
The cases involving $\neg C$ and $C \sqcup D$ can be shown similarly.

**Case** $\langle w{:}\forall R.C \geq n \rangle$**:** Let $\alpha$ be $(w, w'){:}R$ and consider $\langle w'{:}C \geq n \rangle$. It follows that, either $(i)$ $\langle w'{:}C \geq n \rangle \in S'$; or $(ii)$ $\langle w'{:}C \geq n \rangle \notin S'$. Case $(i)$: by induction, $\mathcal{I}$ satisfies $\langle w'{:}C \geq n \rangle$ and, thus, $\max\{1 - R^{\mathcal{I}}(w, w'), C^{\mathcal{I}}(w')\} \geq C^{\mathcal{I}}(w') \geq n$. Case $(ii)$: by construction $R^{\mathcal{I}}(w, w') = glb[(w, w'){:}R, \epsilon[(w, w'){:}R]]$ and $R^{\mathcal{I}}(w, w') \leq 1 - n$ (see Equation 24). It follows that $\max\{1 - R^{\mathcal{I}}(w, w'), C^{\mathcal{I}}(w')\} \geq 1 - R^{\mathcal{I}}(w, w') \geq n$. Therefore, $\inf_{w' \in \Delta^{\mathcal{I}}} \max\{1 - R^{\mathcal{I}}(w, w'), C^{\mathcal{I}}(w')\} \geq n$, *i.e.* $\mathcal{I}$ satisfies $\langle w{:}\forall R.C \geq n \rangle$.

The cases $\langle w{:}\forall R.C > n \rangle$, $\langle w{:}\exists R.C \leq n \rangle$ and $\langle w{:}\exists R.C < n \rangle$ can be shown similarly.

**Case** $\langle w{:}\exists R.C \geq n \rangle$**:** Since $S'$ is complete, both $\langle (w, w'){:}R \geq n \rangle$ and $\langle w'{:}C \geq n \rangle$ are in $S'$. By induction, $\mathcal{I}$ satisfies both $\langle (w, w'){:}R \geq n \rangle$ and $\langle w'{:}C \geq n \rangle$. As a consequence, $\min\{R^{\mathcal{I}}(w, w'), C^{\mathcal{I}}(w')\} \geq n$ follows and, thus, $\sup_{w' \in \Delta^{\mathcal{I}}} \min\{R^{\mathcal{I}}(w, w'), C^{\mathcal{I}}(w')\} \geq n$, *i.e.* $\mathcal{I}$ satisfies $\langle w{:}\exists R.C \geq n \rangle$.

The cases $\langle w{:}\exists R.C > n \rangle$, $\langle w{:}\forall R.C \leq n \rangle$ and $\langle w{:}\forall R.C < n \rangle$ can be shown similarly. $\quad\Box$
The following example shows how such an interpretation is build.

**Example 8** Let us consider Example 6 and the fuzzy assertion $\psi'$. We have shown that $\Sigma \not\models \psi'$ by constructing two clash-free completions from $S$. Let us consider the clash-free completion $S_1$ in branch $\Omega_1$:

$$S_1 = \Sigma \cup \{ \ \langle a{:}\exists R.(D \sqcap C) < 0.5 \rangle, \langle (a, x){:}R \geq 0.7 \rangle, \langle x{:}D \geq 0.7 \rangle, \langle x{:}C \geq 0.4 \rangle,$$
$$\langle b{:}D \sqcap C < 0.5 \rangle, \langle x{:}D \sqcap C < 0.5 \rangle, \langle b{:}D < 0.5 \rangle, \langle x{:}C < 0.5 \rangle \}$$

We show that $S_1$ is satisfiable by building an interpretation as described in the proof of Proposition 3. Accordingly, for $\epsilon_i > 0$,

$$\begin{array}{llll}
glb[(a,b){:}R, \epsilon_1] & = & 0.5 & lub[(a,b){:}R, \epsilon_1] & = & 1 \\
glb[b{:}C, \epsilon_2] & = & 0.2 & lub[b{:}C, \epsilon_2] & = & 1 \\
glb[b{:}D, \epsilon_3] & = & 0.3 & lub[b{:}D, \epsilon_3] & = & 0.5 - \epsilon_3 \\
glb[(a,x){:}R, \epsilon_4] & = & 0.7 & lub[(a,x){:}R, \epsilon_4] & = & 1 \\
glb[x{:}D, \epsilon_5] & = & 0.7 & lub[x{:}D, \epsilon_5] & = & 1 \\
glb[x{:}C, \epsilon_6] & = & 0.4 & lub[x{:}C, \epsilon_6] & = & 0.5 - \epsilon_6.
\end{array}$$

Therefore, we can freely choose $\epsilon_2, \epsilon_4$ and $\epsilon_5$, whereas $\epsilon_3$ and $\epsilon_6$ have to be such that $0.3 \leq 0.5 - \epsilon_3$ and $0.4 \leq 0.5 - \epsilon_6$. As both $\langle(a,b){:}R \geq 0.5\rangle$ and $\langle a{:}\forall R.C \geq 0.4\rangle$ are in $S_1$, while $\langle b{:}C \geq 0.4\rangle \notin S_1$, we have the additional restriction on the choice of $\epsilon_1$ (see Equation 24) that $glb[(a,b){:}R, \epsilon_1] \leq 0.6$. But, $glb[(a,b){:}R, \epsilon_1] = 0.5 \leq 0.6$, for every $\epsilon_1$ and, thus, the choice of $\epsilon_1$ is also free. A solution to the $\epsilon_i$ is $e.g.$ $\epsilon_i = 0.1$ and, thus, let $\mathcal{I}$ be the following interpretation:

1. the domain $\Delta^{\mathcal{I}}$ is the set $\{a, b, x\}$;

2. $w^{\mathcal{I}} = w$, for all $w \in \Delta^{\mathcal{I}}$;

3. $\top^{\mathcal{I}}(w^{\mathcal{I}}) = 1$ and $\bot^{\mathcal{I}}(w^{\mathcal{I}}) = 0$, for all $w \in \Delta^{\mathcal{I}}$;

4. $C^{\mathcal{I}}(b) = 0.2$, $D^{\mathcal{I}}(b) = 0.3$, $D^{\mathcal{I}}(x) = 0.7$ and $C^{\mathcal{I}}(x) = 0.4$ (in all other cases, $A^{\mathcal{I}}(w) = 0$); and

5. $R^{\mathcal{I}}(a,b) = 0.5$ and $R^{\mathcal{I}}(a,x) = 0.7$ (in all other cases, $R'^{\mathcal{I}}(w,w') = 0$).

Now, it is easily verified that $\mathcal{I}$ satisfies $S_1$ and $S$. $\qquad\square$

From a computational complexity point of view, the fuzzy entailment problem can be proven to be a PSPACE-complete problem, as is the classical entailment problem.

**Proposition 4** *Let $\Sigma$ be a fuzzy KB and let $\psi$ be a fuzzy assertion. Determining whether $\Sigma \approx\!\!\!\!\mid \psi$ is a PSPACE-complete problem.* $\qquad\dashv$

*Proof:* We have seen that termination of the above algorithm is guaranteed. Additionally, for a crisp KB $K$, define $\Sigma_K = \{\langle \alpha \geq 1 \rangle : \alpha \in K\}$. By definition, each $\psi \in \Sigma_K$ is KB-normalised and $\langle \alpha \geq 0.5 \rangle$ is query-normalised. Then from Proposition 2 it follows that $K \models \alpha$ iff $\Sigma_K \approx\!\!\!\!\mid \langle \alpha \geq 0.5 \rangle$. From the PSPACE-completeness of the entailment problem in crisp $\mathcal{ALC}$ (Schmidt-Schauß & Smolka, 1991), PSPACE-hardness of the fuzzy entailment problem follows. Unfortunately, our algorithm, as it is, requires exponential space due a well know problem inherited from the crisp case. Indeed, it easily verified that a completion of $S = \{x{:}C\}$, where $C$ is the concept

$$(\exists R.A_{11}) \sqcap (\exists R.A_{12}) \sqcap \forall R.((\exists R.A_{21}) \sqcap (\exists R.A_{22}) \sqcap \ldots \forall R.((\exists R.A_{n1}) \sqcap (\exists R.A_{n2})) \ldots)$$

contains at least $2^n + 1$ variables. In order to require polynomial space, Schmidt-Schauß and Smolka (1991) introduced the so-called *trace rule* $(T\exists)$ for the $\exists$ operator. The $(T\exists)$ rule modifies the $(\exists)$ rule as shown below.

$(\exists) \quad w{:}\exists R.C \rightarrow (w,x){:}R, x{:}C$
> if $x$ new variable and there is no $w'$ such that both $(w,w'){:}R$ and $w'{:}C$
> are already in the actual constraint set

$(T\exists) \quad w{:}\exists R.C \rightarrow (w,x){:}R, x{:}C$
> if $x$ new variable and no $(w,w'){:}R'$ is already in the actual set of constraints.

The difference between the $(\exists)$ rule and the $(T\exists)$ is that the latter is applied only once for an object $w$. We are thus compelled to make a nondeterministic choice amongst the constraints of the form $w{:}\exists R.C$. Furthermore, it is convenient to apply a trace rule only if none of the other $(\sqcap), (\sqcup), (\neg)$ and $(\forall)$ rules are applicable.

We say that a set of constraints $S'$ is a *trace* of a set $S$ if $S'$ obtained from $S$ by application of the rules where the $(\exists)$ has been replaced by $(T\exists)$. Schmidt-Schauß and Smolka (1991) have shown that a set of constraints $S = \{x{:}C\}$ is satisfiable iff no trace $S'$ of $S$ contains a clash. As the size of a trace $S'$ of $S$ is bounded polynomially by the size of $S$, polynomial space is sufficient to prove satisfiability.

The above trace rule works if we start from a constraint set of the form $\{x{:}C\}$. In the general case, we have to rely on so-called *pre-completions* (Baader & Hollunder, 1991b; Donini et al., 1994). A set of constraints $S'$ is said to be a *pre-completion* of a given set of constraints $S$, if it is obtained from $S$ by the application of the $(\sqcap), (\sqcup), (\neg)$ and $(\forall)$ rules, and none of these rules is applicable to $S'$ (the size of $S'$ is polynomially bounded by the size of $S$). As a consequence of this "pre-processing" step, all role relationships $(w,w'){:}R \in S'$ can be ignored, *i.e.* removed from $S'$, because they no longer carry any additional information. Now, in a second step we can apply the method above by checking whether no trace from $S'$ contains a clash. In summary, a set of constraints $S$ is satisfiable iff there is a pre-completion $S'$ of $S$ such that no trace $S''$ of $S'$ contains a clash.

In the fuzzy case, similar trace rules can easily be defined. For instance, the correspondent trace rule of the $(\exists_{\geq})$ rule is

$(T\exists_{\geq}) \quad \langle w{:}\exists R.C \geq n \rangle \rightarrow \langle (w,x){:}R \geq n \rangle, \langle x{:}C \geq n \rangle$
> if $x$ new variable and no $\langle (w,w'){:}R' \geq m \rangle$ is already in
> the actual set of fuzzy constraints.

The trace rules correspondent to the rules $(\exists_{>}), (\forall_{\leq})$ and $(\forall_{<})$ are defined similarly. By proceeding as for the crisp case, it can be shown that $(i)$ a set of fuzzy constraints $S$ is satisfiable iff there is a pre-completion $S'$ of $S$ such that no trace $S''$ of $S'$ contains a clash; and $(ii)$ the size of a trace $S''$ of $S$ is bounded polynomially by the size of $S$. As a consequence, the satisfiability problem is in PSPACE, which completes the proof. $\quad\square$

This result establishes an important property about our fuzzy DL. In effect, it says that no additional computational cost has to be paid for the major expressive power.

## 5.2 A Decision Procedure for the Subsumption Problem

In this section we address the subsumption problem, *i.e.* deciding whether $C \preceq_{\Sigma_T} D$, where $C$ and $D$ are two concepts and $\Sigma$ is a fuzzy terminology. As we have seen (see Example 4), $C \preceq_{\Sigma_T} D$ can be reduced to the case of an empty terminology by applying the KB expansion process. So, without loss of generality, we can limit our attention to the case $C \preceq_\emptyset D$.

At first, an analogue to relation (12) holds. In fact, it can easily be shown that

**Proposition 5** *Let $C$ and $D$ be two concepts. It follows that $C \preceq_\emptyset D$ iff for all $n > 0$ $\langle a{:}C \geq n \rangle \approx \langle a{:}D \geq n \rangle$, where $a$ is a new individual.* ⊣

*Proof:*

⇒ .) Assume that $C \preceq_\emptyset D$ holds. Suppose to the contrary that $\exists n > 0$ such that $\langle a{:}C \geq n \rangle \not\approx \langle a{:}D \geq n \rangle$. Therefore, there is an interpretation $\mathcal{I}$ and an $n > 0$ such that $C^{\mathcal{I}}(a^{\mathcal{I}}) \geq n$ and $D^{\mathcal{I}}(a^{\mathcal{I}}) < n$. But, from the hypothesis $n \leq C^{\mathcal{I}}(a^{\mathcal{I}}) \leq D^{\mathcal{I}}(a^{\mathcal{I}}) < n$ follows. Absurd.

⇐ .) Assume that for all $n > 0$, $\langle a{:}C \geq n \rangle \approx \langle a{:}D \geq n \rangle$ holds. Suppose to the contrary that $C \not\preceq_\emptyset D$ holds. Therefore, there is an interpretation $\mathcal{I}$ and $d \in \Delta^{\mathcal{I}}$ such that $C^{\mathcal{I}}(d) > D^{\mathcal{I}}(d) \geq 0$. Let us extent $\mathcal{I}$ to $a$ such that $a^{\mathcal{I}} = d$ and consider $\overline{n} = C^{\mathcal{I}}(d) > 0$. Of course, $\mathcal{I}$ satisfies $\langle a{:}C \geq \overline{n} \rangle$. Therefore, from the hypothesis it follows that $\mathcal{I}$ satisfies $\langle a{:}D \geq \overline{n} \rangle$, *i.e.* $D^{\mathcal{I}}(d) \geq \overline{n} = C^{\mathcal{I}}(d) > D^{\mathcal{I}}(d)$. Absurd. □

How can we check whether for all $n > 0$ $\langle a{:}C \geq n \rangle \approx \langle a{:}D \geq n \rangle$ holds? A solution to this problem, restricted to the propositional case, is given by Straccia (2000a). Indeed, it is shown that

**Proposition 6 (Straccia, 2000a)** *Let $p$ and $q$ be two propositions, $0 < n_1 \leq 0.5$ and $1 \geq n_2 > 0.5$. It follows that for all $n > 0$, $\langle p \geq n \rangle \approx \langle q \geq n \rangle$ iff for both $m \in \{n_1, n_2\}$, $\langle p \geq m \rangle \approx \langle q \geq m \rangle$ holds.* ⊣

The above proposition establishes that, at the propositional level, in order to check whether $\langle p \geq n \rangle \approx \langle q \geq n \rangle$ holds for all $n$, it is sufficient to check the entailment relation with respect to two values $n_1, n_2$. The first being less or equal than 0.5, while the second being greater than 0.5, respectively. This is due to the fact that for given values $n, n' \leq 0.5$, any proof of $\langle p \geq n \rangle \approx \langle q \geq n \rangle$ can be converted into a proof for $\langle p \geq n' \rangle \approx \langle q \geq n' \rangle$ and vice-versa. The case where $n, n' > 0.5$ is similar.

The above proposition can be extended to our fuzzy DL as well.

**Lemma 1** *Let $C$ and $D$ be two concepts, $0 < n, n' \leq 0.5$ and let $a$ be an individual. It follows that $\langle a{:}C \geq n \rangle \approx \langle a{:}D \geq n \rangle$ iff $\langle a{:}C \geq n' \rangle \approx \langle a{:}D \geq n' \rangle$.* ⊣

*Proof:* It is enough to show that $S = \{\langle a{:}C \geq n \rangle, \langle a{:}D < n \rangle\}$ is satisfiable iff $S' = \{\langle a{:}C \geq n' \rangle, \langle a{:}D < n' \rangle\}$ is satisfiable.

⇒ .) Assume that $S$ is satisfiable. So, there is a clash-free completion $\tilde{S}$ of $S$. With $\tilde{S}[n/n']$ we indicate the set of fuzzy constraints obtained from $\tilde{S}$, by replacing any value $n$ in $\tilde{S}$ with $n'$ and any value $1 - n$ in $\tilde{S}$ with $1 - n'$, respectively. We will show that $\tilde{S}[n/n']$ is a clash-free completion of $S'$ and, thus, $S'$ is satisfiable.

Let $r_1, \ldots, r_k$, $k \geq 0$ be the sequence of inference rule applications, which applied to $S$ get $\tilde{S}$. Let $\tilde{S}_0 = S$, let $\tilde{S}_k = \tilde{S}$ and for $1 \leq i \leq k$ let $\tilde{S}_i$ be the set of fuzzy constraints obtained from $\tilde{S}_{i-1}$ by the application of the $r_i$ rule to $\tilde{S}_{i-1}$.

By induction on $k$, we show that $(i)$ the sequence of inference rules $r_1, \ldots, r_k$ can be applied to $S'$ as well; $(ii)$ for $\tilde{S}'_0 = S'$, $\tilde{S}'_k = \tilde{S}'$ and $\tilde{S}'_i$ the set of fuzzy constraints obtained from $\tilde{S}'_{i-1}$ by the application of the $r_i$ rule to $\tilde{S}'_{i-1}$, we have that $\tilde{S}'_i = \tilde{S}_i[n/n']$ and, thus, $\tilde{S}' = \tilde{S}'_k = \tilde{S}_k[n/n'] = \tilde{S}[n/n']$; and $(iii)$ if $\tilde{S}_k$ is a clash-free completion of $S$ then $\tilde{S}'_k$ is a clash-free completion of $S'$ as well.

**case** $k = 0$**:** No rule is applicable to $S$ and $\tilde{S}_0 = S$ is a completion of $S$. By case analysis, $(i)$ it is easily verified that no rule is applicable to $S'$ and $(ii)$ $S' = \tilde{S}'_0 = \tilde{S}_0[n/n'] = S[n/n']$. $(iii)$ So, $\tilde{S}'_0$ is a completion of $S'$. We show that $\tilde{S}'_0$ is clash-free. Assume to the contrary that $\tilde{S}'_0$, $i.e.$ $S'$ contains a clash. As a consequence, $S' = \{\langle a{:}C \geq n'\rangle, \langle a{:}C < n'\rangle\}$. But then, $S$ is $\{\langle a{:}C \geq n\rangle, \langle a{:}C < n\rangle\}$, contrary to the assumption that $S$ is clash-free.

**induction step:** by case analysis on the rule $r_k$. We limit our presentation to the $(\forall_\geq)$ rule as for the other the proof is similar.

If $r_k$ is $(\forall_\geq)$ then it can be verified that there are $\langle w{:}\forall R.C' \geq n\rangle$ and $\langle (w, w'){:}R > 1 - n\rangle$ in $\tilde{S}_{k-1}$ such that $\langle w'{:}C' \geq n\rangle \notin \tilde{S}_{k-1}$ and $\langle w'{:}C' \geq n\rangle \in \tilde{S}_k$. By induction, $\tilde{S}_{k-1}[n/n'] = \tilde{S}'_{k-1}$ and, thus, both $\langle w{:}\forall R.C' \geq n'\rangle$ and $\langle (w, w'){:}R > 1 - n'\rangle$ are in $\tilde{S}'_{k-1}$, while $\langle w'{:}C' \geq n'\rangle \notin \tilde{S}'_{k-1}$. Therefore, $(i)$ rule $r_k$ is applicable to $\tilde{S}'_{k-1}$ and $\langle w'{:}C' \geq n'\rangle \in \tilde{S}'_k$; $(ii)$ so, $\tilde{S}'_k = \tilde{S}_k[n/n']$; $(iii)$ from $\tilde{S}'_k = \tilde{S}_k[n/n']$ and, as $\tilde{S}_k$ is a completion of $S$, by case analysis, it is easily verified that no rule is further applicable to $\tilde{S}'_k$. Therefore, $\tilde{S}'_k$ is a completion of $S'$. Let us show that $\tilde{S}'_k$ is clash free. Assume to the contrary that $\tilde{S}'_k$ contains a clash. If one of the cases in Table 1 holds, then from $\tilde{S}'_k = \tilde{S}_k[n/n']$ it follows easily that there is a clash in $\tilde{S}_k$ as well, which is contrary to assumption that $\tilde{S}_k$ is clash-free. On the other hand, if there is a conjugated pair of fuzzy constraints in $\tilde{S}'_k$ (see Table 2), then one of the following three pairs is in $\tilde{S}'_k$: (a) $\langle \alpha \geq n'\rangle$ and $\langle \alpha < n'\rangle$; (b) $\langle \alpha \leq 1 - n'\rangle$ and $\langle \alpha > 1 - n'\rangle$; and (c) $\langle \alpha < n'\rangle$ and $\langle \alpha > 1 - n'\rangle$ (note that $n, n' \leq 0.5$). Again, as $\tilde{S}'_k = \tilde{S}_k[n/n']$, it follows that there is a conjugated pair in $\tilde{S}_k$ as well, which is contrary to assumption that $\tilde{S}_k$ is clash-free.

$\Leftarrow$ .) Can be proven similarly to $\Rightarrow$ .). $\quad \square$

By proceeding as for Lemma 1 it can be shown that

**Lemma 2** *Let $C$ and $D$ be two concepts, $1 \geq n, n' > 0.5$ and let $a$ be an individual. It follows that $\langle a{:}C \geq n\rangle \!\approx\!\langle a{:}D \geq n\rangle$ iff $\langle a{:}C \geq n'\rangle \!\approx\!\langle a{:}D \geq n'\rangle$.* $\quad \dashv$

From Lemma 1 and Lemma 2 it follows that

**Proposition 7** *Let $C$ and $D$ be two concepts, $0 < n_1 \leq 0.5$, $1 \geq n_2 > 0.5$ and let $a$ be an individual. It follows that for all $n > 0$ $\langle a{:}C \geq n\rangle \!\approx\!\langle a{:}D \geq n\rangle$ iff for both $m \in \{n_1, n_2\}$, $\langle a{:}C \geq m\rangle \!\approx\!\langle a{:}D \geq m\rangle$ holds.* $\quad \dashv$

As a consequence, the subsumption problem can be reduced to the entailment problem for which we have a decision algorithm.

### 5.3 A Decision Procedure for the BTVB Problem

We address now the problem of determining $glb(\Sigma, \alpha)$ and $lub(\Sigma, \alpha)$. This is important, as computing, *e.g.* $glb(\Sigma, \alpha)$, is in fact the way to answer a query of type "to which degree is $\alpha$ (at least) true, given the (imprecise) facts in $\Sigma$?".

Without loss of generality, we will assume that all concepts are in NNF (*Negation Normal Form*). Straccia (2000b) has shown that, in case of fuzzy propositional logic, from a set $\Sigma$ of fuzzy propositions of the form $\langle p \geq n \rangle$ and $\langle p \leq n \rangle$, where $p$ is a proposition, it is possible to determine a finite set $N^{\Sigma} \subset [0, 1]$, where $|N^{\Sigma}|$ is $O(|\Sigma|)$, such that $glb(\Sigma, q) \in N^{\Sigma}$, *i.e.* the greatest lower bound of a proposition $q$ w.r.t. $\Sigma$ has to be an element of $N^{\Sigma}$. Therefore, $glb(\Sigma, q)$ can be determined by computing the greatest value $n \in N^{\Sigma}$ such that $\Sigma \not\models \langle q \geq n \rangle$. An easy way to search for this $n$ is to order the elements of $N^{\Sigma}$ and then to perform a binary search among these values by successive entailment tests. Dually, as $lub(\Sigma, q) = 1 - glb(\Sigma, \neg q)$ holds, the $lub$ can either be computed from the $glb$ or, as $lub(\Sigma, q) \in 1 - N^{\Sigma}$, where $1 - N^{\Sigma} = \{1 - n : n \in N^{\Sigma}\}$, we can compute it by determining the smallest value in $1 - N^{\Sigma}$.

**Proposition 8 (Straccia, 2000b)** *Let $\Sigma$ be a set of fuzzy propositions in NNF and let $q$ be a proposition. Then $glb(\Sigma, q) \in N^{\Sigma}$ and $lub(\Sigma, q) \in 1 - N^{\Sigma}$, where*

$$
\begin{aligned}
N^{\Sigma} &= \{0, 0.5, 1\} \ \cup \\
&\quad \{n : \langle p \geq n \rangle \in \Sigma\} \ \cup \\
&\quad \{1 - n : \langle p \leq n \rangle \in \Sigma\} \\[2mm]
1 - N^{\Sigma} &= \{1 - n : n \in N^{\Sigma}\}.
\end{aligned}
$$

⊣

The above Proposition 8 can easily be extended our fuzzy description logic case. Essentially, the quantifiers do not change the possible values of $glb(\Sigma, F)$ and $lub(\Sigma, F)$.

**Proposition 9** *Let $\Sigma$ be a set of fuzzy assertions in NNF and let $\alpha$ be an assertion. Then $glb(\Sigma, \alpha) \in N^{\Sigma}$ and $lub(\Sigma, \alpha) \in 1 - N^{\Sigma}$, where*

$$
\begin{aligned}
N^{\Sigma} &= \{0, 0.5, 1\} \ \cup \\
&\quad \{n : \langle \alpha \geq n \rangle \in \Sigma\} \ \cup \\
&\quad \{1 - n : \langle \alpha \leq n \rangle \in \Sigma\} \\[2mm]
1 - N^{\Sigma} &= \{1 - n : n \in N^{\Sigma}\}.
\end{aligned}
$$

⊣

*Proof:* Let us show that $glb(\Sigma, \alpha) \in N^{\Sigma}$. Let $m$ be $glb(\Sigma, \alpha)$. By definition, if $m = 0$ then $S = \Sigma \cup \{\langle \alpha < n \rangle\}$ is satisfiable for any $n > 0$ and $0 \in N^{\Sigma}$. Otherwise, $m > 0$ is the largest value such that $S = \Sigma \cup \{\langle \alpha < m \rangle\}$ is not satisfiable. Let us mark each sub-expression in $\alpha$ with a *, so that we can trace the components of the query assertion $\alpha$ during a deduction.

Consider a completion $S'$ of $S$. Starting from $\langle \alpha < m \rangle$, by applying the rules of inference, only * marked expressions of type $\langle \alpha' < m \rangle$ or $\langle \alpha' > 1 - m \rangle$ can appear in $S'$. Furthermore, as $S$ is not satisfiable, $S'$ contains a clash, *i.e.* the value $m$ is the largest value such that all completions $S'$ of $S$ contain a clash. Let us analyse $S'$. As $S'$ contains a clash, then either $(i)$ there is a clash according to Table 1, or $(ii)$ there is a clash according to Table 2. If $(i)$ is the case, *i.e.* there is $\psi \in S'$ which is a clash, then we have to distinguish between two cases: (a) $\psi$ is not marked with * and (b) $\psi$ is marked with *. In the former case, $S'$ contains a clash independently of the value $m$ and, thus, the largest possible value $m$ for which $S'$ contains a clash is 1. In the latter case, as $m > 0$, either $\psi = \langle w{:}\bot > 1 - m \rangle$ or $\psi = \langle w{:}\top < m \rangle$ which are both clashes for any value of $m$. As a consequence, the largest possible value $m$ for which $S'$ contains a clash according to Table 1 is 1. Assume $(ii)$ is the case, *i.e.* a conjugated pair of fuzzy constraints $\psi$ and $\psi'$ is in $S'$. Similarly to the previous case, we have to distinguish the cases for which $\psi$ and $\psi'$ are marked with *. There are four cases:

(a) $\psi = \langle \alpha \geq k \rangle$ and $\psi' = \langle \alpha' \leq k' \rangle$ are in $S'$, none is marked with * and $k > k'$. Therefore, $S'$ contains a clash for any value of $m$ and, thus, the largest choice is 1;

(b) $\psi = \langle \alpha \geq k \rangle$ and $\psi' = \langle \alpha' < m \rangle$ are in $S'$, only $\psi'$ is marked with * and $k \geq m$. Therefore, $S'$ contains a clash for any value of $m \leq k$ and, thus, the largest choice for $m$ is $k$. It is easily verified by case analysis on the rules that from $\psi \in S'$, $k \in N^{\Sigma}$ follows;

(c) $\psi = \langle \alpha \leq k \rangle$ and $\psi' = \langle \alpha' > 1 - m \rangle$ are in $S'$, only $\psi'$ is marked with * and $k \leq 1 - m$. Therefore, $S'$ contains a clash for any value of $m \leq 1 - k$ and, thus, the largest choice for $m$ is $1 - k$. It is easily verified by case analysis on the rules that from $\psi \in S'$, $1 - k \in N^{\Sigma}$ follows;

(d) $\psi = \langle \alpha < m \rangle$ and $\psi' = \langle \alpha' > 1 - m \rangle$ are in $S'$, both are marked with * and $m \leq 1 - m$. Therefore, $S'$ contains a clash for any value of $m \leq 0.5$ and, thus, the largest choice for $m$ is 0.5.

Summing up, we have proved that the largest possible value for $m$ is such that $m \in N^{\Sigma}$. $\quad \square$

The algorithms computing $glb(\Sigma, \alpha)$ and $lub(\Sigma, \alpha)$ are described in Table 3. For instance, by a binary search on $N^{\Sigma}$, the value of $glb(\Sigma, \alpha)$ can be determined in at most $\log |N^{\Sigma}|$ fuzzy entailment tests.

# 6. Conclusions and Future Work

In this work, we have presented a quite general fuzzy extension of the DL $\mathcal{ALC}$, a significant and expressive representative of the various DLs. Our fuzzy DL enables us to reason in presence of imprecise $\mathcal{ALC}$ concepts, *i.e.* fuzzy $\mathcal{ALC}$ concepts. From a semantics point of view, fuzzy concepts are interpreted as fuzzy sets *i.e.* given a concept $C$ and an individual $a$, $C(a)$ is interpreted as the truth-value of the sentence "$a$ is $C$". From a syntax point of view, we allow to specify lower and upper bounds of the truth-value of $C(a)$. Complete algorithms for reasoning in it have been presented, that is, we have devised algorithms for solving the entailment problem, the subsumption problem as well as the best truth-value bound problem.

---

**Algorithm** $Max(\Sigma, \alpha)$
Set $Min := 0$ and $Max := 2$.

1. Pick $n \in N^\Sigma \setminus \{0\}$ such that $Min < n < Max$. If there is no such $n$, then set $glb(\Sigma, \alpha) := Min$ and exit.

2. If $\Sigma \not\models \langle \alpha \geq n \rangle$ then set $Min = n$, else set $Max = n$. Go to Step 1.

**Algorithm** $Min(\Sigma, \alpha)$
Set $Min := 0$ and $Max := 2$.

1. Pick $n \in (1 - N^\Sigma) \setminus \{0\}$ such that $Min < n < Max$. If there is no such $n$, then set $lub(\Sigma, \alpha) := \min\{Max, 1\}$ and exit.

2. If $\Sigma \not\models \langle \alpha \leq n \rangle$ then set $Max = n$, else set $Min = n$. Go to Step 1.

---

Table 3: Algorithms $Max(\Sigma, \alpha)$ and $Min(\Sigma, \alpha)$

An important point concerns computational complexity. The complexity result shows that the additional expressive power has no impact from a computational complexity point of view.

The extension of DLs to the management of vagueness is not new (Tresp & Molitor, 1998; Yen, 1991). Yen was the first, to the best of my knowledge, introducing vagueness into a simple DL. His language has two interesting points not included into our language. Firstly, it allows the definition of vague concepts by means of explicit membership functions over a domain, *e.g.* LowPressure:$\approx domain$(AirPressure); $membershipfx(\lambda p.\mathsf{low}(p))$. Here the domain over $p$ ranges is given by AirPressure. $\lambda p.\mathsf{low}(p)$ determines the membership degree of being a pressure $p$ low. Secondly, the language allows *concept modifiers*, like Very or Slightly, by means of which concepts like "very low pressure" can be defined through VeryLowPressure:$\approx$ Very(LowPressure). This last idea has been generalised to $\mathcal{ALC}$ by Tresp and Molitor (1998) where a certain type of concept modifiers are allowed. Strictly speaking, the language defined by Tresp and Molitor is more expressive, as we do not consider concept modifiers. From a semantics point of view, the extension to Tresp and Molitor's language is quite straightforward. But, the cost that we have to pay for this increasing expressive power is that, from a computational complexity and algorithms point of view, things changes radically. Indeed, according to Tresp and Molitor for each "completion" a linear optimisation problem is generated (set of inequations of the form $op_1(t_1)Rop_2(t_2)$ or $op_1(t_1)Rf(op_2(t_2))$, where $t_i$ is a truth-value variable, $R \in \{\leq, \geq, =, op_i \in \{id, m_i\}$, with $id(t)$ the identity and $m_i(t)$ is a modifier function over truth-value variables, respectively, and $f \in \{min, max\}$-derived from the semantics of the connectors $\sqcap, \sqcup$) and then solved for the best value. Then, the minimum among all computed solutions is taken. The solutions can be computed by relying on methods from the domain of linear programming, *e.g.* the *simplex method* (Papadimitriou & Steiglitz, 1982). While it is possible to devise a similar approach for our fuzzy DL as well, we have seen that this is not necessary.

Both aspects considered by Tresp and Molitor and by Yen, although interesting, are not crucial w.r.t. how we model logic-based multimedia information retrieval, where underlaying text, image and video analysis tools provide us fuzzy assertions *e.g.* of the form $\langle a{:}\exists\mathsf{About}.C \geq n\rangle$ from which we may infer $\langle a{:}\exists\mathsf{About}.D \geq m\rangle$.

This work can be used as a basis both for extending existing DL based systems and for further research. In this latter case, there are several open points. For instance, it is not clear yet how to reason both in case of fuzzy specialisation of the general form $C \prec D$ and in the case cycles are allowed in a fuzzy KB. Another interesting topic for further research concerns the semantics of fuzzy connectives. Of course several other choices for the semantics of the connectives $\sqcap, \sqcup, \neg, \exists, \forall$ can be considered. While for a huge number of proposals given in the literature their impact from a semantics point of view is well understood, the question how they impact from a computational complexity and algorithms point of view remains still open.

## Acknowledgements

I would like to thank the three anonymous reviewers for their helpful comments on an early version of this paper. This is an extension and revision of the paper appeared in AAAI-98.